\documentclass{elbioimp2}

\usepackage{enumitem}
\usepackage[utf8]{inputenc}
\usepackage{geometry}
\usepackage{array}
\usepackage[backend=biber,style=vancouver]{biblatex}
\usepackage{csquotes}
\usepackage{amsmath}
\usepackage{tikz}
\usepackage{listings}
\usepackage{graphicx}
\usepackage{animate}
\usepackage{hyperref}
\usepackage{tabularx}
\usepackage{booktabs}
\usepackage{tikz}
\usepackage[linesnumbered,ruled,vlined]{algorithm2e}

\usetikzlibrary{shapes, arrows, positioning}

\usetikzlibrary{shapes,arrows}
\usetikzlibrary{positioning}
\title{Optimizing Mario Adventures in a Constrained Environment}
\shorttitle{Mario GA}
\author{Sanyam Jain\affiliation{Østfold University College, Halden, Norway, sanyamj@hiof.no}} 
\shortauthor{Jain et al.}
 
\begin{document}
\maketitle

\begin{abstract}
  This project proposes and compares a new way to optimise Super Mario Bros. (SMB) environment where the control is in hand of two approaches, namely, Genetic Algorithm (MarioGA) and Neuro-Evolution (MarioNE). Not only we learn playing SMB using these techniques, but also optimise it with constrains of collection of coins and finishing levels. Firstly, we formalise the SMB agent to maximize the total value of collected coins (reward) and maximising the total distance traveled (reward) in order to finish the level faster (time penalty) for both the algorithms. Secondly, we study MarioGA and its evaluation function (fitness criteria) including its representation methods, crossover used, mutation operator formalism, selection method used, MarioGA loop, and few other parameters. Thirdly, MarioNE is applied on SMB where a population of ANNs with random weights is generated, and these networks control Mario's actions in the game. Fourth, SMB is further constrained to complete the task within the specified time, rebirths (deaths) within the limit, and performs actions or moves within the maximum allowed moves, while seeking to maximize the total coin value collected. This ensures an efficient way of finishing SMB levels. Finally, we provide a five-fold comparative analysis by plotting fitness plots, ability to finish different levels of world 1, and domain adaptation (transfer learning) of the trained models. Overall, MarioGA was able to finish levels 1-1,1-2,1-3 and 1-4 94\% times where as MarioNE was rarely able to finish with 8\% times with average fitness of 3000+ and 2000+ respectively. This experimentation is five fold ran for 36 hours, where MarioGA was able to perform 8000 generations (if not converge before) while MarioNE was able to perform approx 700 generations. Overall, this problem provides a shortest path that a Mario agent should take in order to finish level. We release all code here - \url{https://github.com/s4nyam/MarioEC}

\end{abstract}

\section{Introduction}
In agent-based systems (for example, Robotics) it is critical to have an efficient agent which can preform dedicated tasks in constrained environment optimally. To formalise such tasks in a simulated environment, we choose "gym-super-mario-bros" library \cite{gym-super-mario-bros} which is an open-source Python package that extends the Gym framework to provide a framework for experimenting with SMB. In order to achieve a task-based control in such constrained environment or dealing with this constrained optimisation problem (COP) we propose MarioGA and MarioNE which implements and evaluates the SMB environment such that our agent (Mario) could complete levels within time limits and collecting as many coins along with distance covered. We further discuss lot more details why we hypothesise such fitness function in following sections.

A detailed SMB environment variables are presented in Table \ref{tab:1} SMB environment has been studied a lot in different setups recently \cite{mautschke2019evolutionary,bytycci2020combination}. For example in \cite{jain2023ramario} has provided a experimental study of Proximal Policy Optimization (PPO), Deep Q-Network (DQN), and their proposed approach of Reptile algorithm (a.k.a RAMario). However, our focus restricts more towards Genetic Algorithm (MarioGA) and Neuro-Evolution (MarioNE). We present and study more about previous works in GA and NE in next section.

Based on the availability of input, output and model, there is a classification for Optimisation, Modelling and Simulation in \cite{eiben2015introduction}, our problem fits with Optimisation. Additionally, we also know this is a COP because we have output specified as the best path that Mario traverses such that it is travels more and collects maximum coins in its adventure, keeping with constraints of time, actions or movement done, etc. So the output is specified however we do not know exactly what input combinations will give this. Further, for a given instance of environment, we can always know (using a formula or calculations) how many coins it has collected so far at the moment, distance traveled, movements or actions performed, number of rebirths, and time elapsed so far at the moment. Hence model is also defined. We formulate a COP over state space $S$, defining $\psi(s)$ as the constraint function, ensuring satisfaction of specified constraints, and introducing a new function $g(s)$ that quantifies constraint violations; thus, a game configuration represents a solution to the Mario COP only when $g(s) = 0$ and $\phi(s) = true$, signifying successful coin collection, distance maximization in minimum game time, and constraint adherence. In addition, the problem falls under constrained optimization and is classified within NP due to its efficient verification process for valid solutions, making it suitable for NP algorithms like Evolutionary Computation, particularly when verifying if Mario's path adheres to constraints while collecting at least X coins.

\begin{table}[h]
\centering
\caption{Mario Game Dictionary}
\label{tab:1} 
\begin{tabular}{|c|c|p{5cm}|}
\hline
\textbf{Key} & \textbf{Type} & \textbf{Description} \\
\hline
coins & int & The number of collected coins \\
flag\_get & bool & True if Mario reached a flag or ax \\
life & int & The number of lives left, i.e., \{3, 2, 1\} \\
score & int & The cumulative in-game score \\
stage & int & The current stage, i.e., \{1, ..., 4\} \\
status & str & Mario's status, i.e., \{'small', 'tall', 'fireball'\} \\
time & int & The time left on the clock \\
world & int & The current world, i.e., \{1, ..., 8\} \\
x\_pos & int & Mario's x position in the stage (from the left) \\
y\_pos & int & Mario's y position in the stage (from the bottom) \\
\hline
\end{tabular}
\end{table}

MarioGA uses standard genetic algorithm described in the textbook "Introduction to Evolutionary Computing" \cite{eiben2015introduction}. A standard MarioGA approach is inspired from nature or natural evolution such that we have an environment (SMB) which is filled with candidate solutions (population) that strive for survival and their reproduction (crossover) where their quality (fitness) is determined by environment which provides a criteria whether they (elites) or their children (crossover and mutation) are going to grow in the environment which are fighting for resources (as metaphor) most effectively. Two key terminologies, Genotype and Phenotype (also discussed later in further sections) are core to any individual in the process, where for us in SMB world,  actions, movements (when to jump, move left, move right, and use power-ups), and other Mario's game-play strategies can be classified as genotype where as the resultant behaviour of those encoded action plan is phenotype. To name a few phenotype, it can be, Mario navigating the level, jumping over obstacles, collecting coins, defeating enemies, and completing the level. Here it is important to note that, one possible value of genotype is allele (for example, In this case, the gene associated with Mario's jumping behavior has two possible alleles: "High Jump" and "Low Jump."). An important point to note in this discussion is that phenotypic differences always stem from variances in the genotype, which can further be simplified that genotype encodes the material which abstracts the information necessary for a particular phenotype. We envision to also use crossover which is an operation mimics genetic recombination inspired by meiosis, where genetic information from two parent solutions is combined to produce offspring solutions with diverse traits. The reason or motivation why we propose MarioGA (Evolutionary Algorithm or Genetic Algorithm) is because of the complexity of the problem to be solved. Since in this COP, we do not have much options to tailor or handcraft, which makes it impossible to try out all possible combinations of the game configurations to be used. Since, EAs or GAs are proven to be robust in such cases to provide near-optimal or optimal solutions as given examples in \cite{eiben2015introduction}.

\begin{figure}[htp]
  \centering
  \includegraphics[width=\columnwidth]{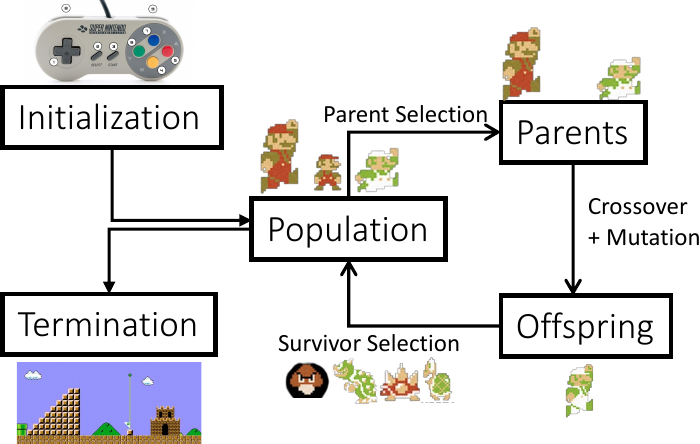}
  \caption{Standard Genetic Algorithm\label{fig:ea}}
\end{figure}
MarioNE uses standard Neuro-Evolution approach described in \cite{floreano2008neuroevolution}. MarioNE utilizes artificial neural networks (ANNs) to enable Mario to learn and adapt his gameplay strategies. Think of it as teaching Mario to become a better player by allowing him to practice and evolve within the game. This approach draws inspiration from nature, where animals strive to survive and reproduce, gradually improving their abilities. MarioNE aims to optimize how Mario plays the game, enhancing his skills in actions like jumping, coin collection, and level completion. It is much simpler and relaxed algorithm comparing to EA or GA such that NE only requires a measure of a network's performance in a given task. For instance, it can evaluate the outcome of a game, such as whether Mario won or lost, without the need for labeled examples of desired strategies \cite{risi2015neuroevolution}.

In following text and further sections, we study Related Works, Proposed Methods (MarioGA and MarioNE), Implementation Details, Results, Discussion and finally Conclusion. Additionally, we study Initialisation, Representation of individuals, Evaluation or Objective function, Population, Selection criteria, Variation operations (Crossover and Mutation) and finally Termination criteria as part of explanation for MarioGA as shown in Figure \ref{fig:ea}. A very similar pipeline is followed by MarioNE however, rather than initialising agent that can perform various Mario operations (configuration), we initialise ANNs that controls SMB and fitness is evaluated on the network's performance on evaluation criteria.

\section{Related Work}
We study two work that are focused in utilising GA and NE to play SMB respectively. Among the approaches in \cite{mautschke2019evolutionary,bytycci2020combination,jain2023ramario,risi2015neuroevolution}, GA leverages evolutionary principles to optimize SMB's gameplay sequences, evolving action plans over generations. On the other hand, NE employs artificial neural networks to evolve intelligent controllers for Mario, allowing him to adapt and learn in dynamic environments. 

\begin{figure}[htp]
  \centering
  \includegraphics[width=0.8\columnwidth]{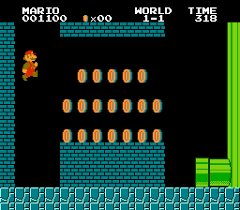}
  \caption{Snapshot from World 1 Level 1 \label{fig:pipe}}
\end{figure}

GAs optimize SMB's gameplay by evolving action sequences over generations, representing potential paths that undergo mutation and crossover for enhanced performance. Fitness evaluation considers coin collection, distance traveled, and constraints, ultimately guiding Mario to collect more coins efficiently in constrained game environments. GAs simulate natural selection, starting with random sequences and selecting the best for reproduction. Over time, these sequences improve, optimizing SMB's gameplay strategy \cite{tonder2013multi}. They propose a very good setup, however they consider fitness as $Distance + (20 * Coins)$ which makes it very ineffective for an agent to travel longer distances. For example, consider for example, a situation in the game where there is a place with a pipe and collection of coins (as shown in Figure \ref{fig:pipe}), this will make our GA loop stuck in the same loop everytime as this fitness is rewarding coins too much. Authors in (section 6.3 of \cite{tonder2013multi})  also mention that this function is prone to stuck in local maximas and where the different objectives will have to be scaled properly in the fitness function in order for it to have any effect. And hence we aim to propose a robust and efficient fitness function.

NE focuses on evolving artificial neural networks (ANNs) as controllers for Mario. It begins with a population of ANNs with random weights. These networks control Mario's actions and are evaluated based on their ability to help Mario achieve goals like coin collection and obstacle avoidance. Successful ANNs are preserved and mutated, gradually improving Mario's decision-making abilities through neural network evolution. NE adapts Mario's behavior to various challenges \cite{mautschke2019evolutionary,bytycci2020combination,risi2015neuroevolution}, making it a valuable tool for optimizing gameplay. In the literature none of the work actually focuses on constrained SMB environment. All of the recent work focuses a lot on how Mario as agent can perceive more elements from environment instead focusing on finishing levels with maximum coin collection in less moves and shorter duration. Moreover, the architectural inefficiency makes it under-performing as previous works are exploring multiple SMB agents that can finish game which increase their runtime of the algorithm. We claim that there is no need to converge for multiple agents, instead we only need one agent that can finish SMB environment level. Authors in \cite{bytycci2020combination} also mention that their limitation is lack of high processing power that would lead also to conduction of more experiments and probably better results. And as part of future work they also mention an approach with a harder version of the game which consists of more constraints can be developed on which our research work in this paper thrives.

We also study deep learning reward based approaches, however compaing MarioGA and MarioNE with such approaches is not scope of this paper. RL methodologies, such as Q-learning and Deep Q-Networks (DQNs), have garnered substantial attention due to their data-driven approach, enabling Mario to adapt and learn optimal strategies through interactions with the game environment. These RL-based methods equip Mario with the ability to make decisions based on rewards and penalties, which proves highly effective in navigating dynamic and evolving in-game scenarios. This adaptability, coupled with the capacity to discover novel gameplay strategies over time, positions RL as a vital component in the quest to optimize Mario's performance in SMB \cite{engelsvoll2020generating}.

While these methods have shown significant promise in enhancing SMB's performance within constrained game environments, however, do not achieve optimal solutions through evolutionary and neural network-based strategies. Our specific constrained and objective is to collect as many coins with more distance traveled while in as short gameplay. We envision that these constraints will help our agent to converge faster and should result a robust SMB agent.
\begin{figure*}[htp]
  \centering
  \includegraphics[width=2\columnwidth]{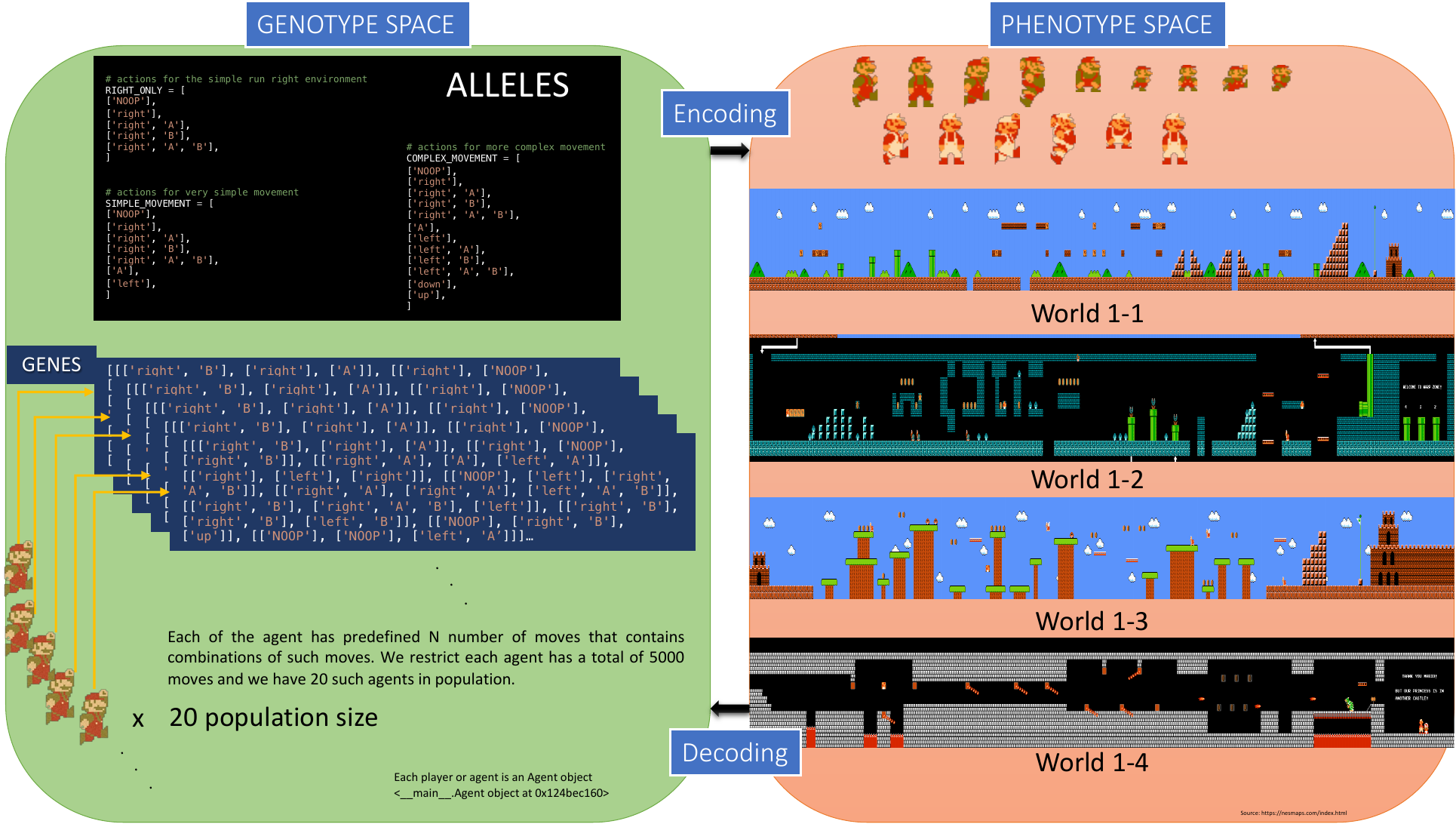}
  \caption{Individual's representation of MarioGA \label{fig:representation}}
\end{figure*}

\section{Proposed Methodology of MarioGA}
In this proposed methodology, we elucidate how individuals, forming a population collectively, are represented. Each individual possesses a fitness or evaluation criteria, and various variation operators, including crossover and mutation, are employed. Selection is driven by higher fitness levels, portraying the principles of "survival of the fittest" and "mating of the fittest," ultimately optimizing on a fitness landscape. We also undertake the consideration of novelty vs quality tradeoff. In this tradeoff, two contrasting dynamics are at play: one aims to boost population variety through genetic mechanisms such as mutation and recombination, promoting the exploration of new possibilities, while the other strives to diminish population variety through selection processes that encompass both parental selection and survivor selection, with the primary goal of enhancing overall quality by exploitation. Additionally, literature discusses Binary, Integer, Real-Valued or Floating-Point, Permutation, and Tree representations for Genetic Operators in \cite{eiben2015introduction}. However, not necessarily the best ones for some applications.

\subsection{Representation}
One very important decision to make in beginning is for representation. As discussed concepts based on phenotype (object) and genotype (code) earlier in introduction, it is wise to note that representation provides a scope for perturbing and modifications using variation operators. We represent MarioGA agents in genotype space using several variables. In our case, the representation takes the form of a series of discrete actions that Mario can perform during the game, including jumping, moving left or right, and using power-ups. Each individual within the population is represented by a unique sequence of these actions, effectively defining its genotype. These sequences serve as the digital DNA of the agents and determine their behavior in the game world. This representation implies two essential mappings: encoding and decoding. Encoding translates the desired in-game behaviors (phenotype) into a corresponding sequence of actions (genotype), which is not necessarily one-to-one. Conversely, decoding translates the genotype back into the in-game behaviors, ensuring a one-to-one correspondence between the genotype and phenotype. These mappings are essential for translating between the genetic encoding and the agent's actual behavior. Furthermore, the representation involves the concept of "genes" and "alleles." Genes correspond to specific elements within the action sequences, while alleles represent the different possible values or actions that a gene can take. For instance, a gene associated with Mario's jumping behavior may have two possible alleles: "UP" and "RIGHT" In this genetic representation, every feasible solution is encoded within the genotype space, allowing the genetic algorithm to evolve and optimize Mario's gameplay strategies over successive generations. 

To be specific about methodology, Agents are initialized with a sequence of moves that serve as their genetic representation as shown in Figure \ref{fig:representation}. This representation is encapsulated within the $self.moves$ attribute of the $Agent$ class, defining the agent's behavior in the game. The $randomize\_moves()$ method in the $Agent$ generates a random sequence of actions to initialize the agent's moves. This is an essential step in creating the genetic representation for a new agent. During gameplay, the agent's actions (genetic representation) are executed and recorded. This recorded sequence of moves represents how the agent performs in the game environment (phenotype). When creating a population of agents, their genetic representations (sequences of moves) are generated using randomization. This initial representation sets the foundation for evolution in subsequent generations.

\subsection{Evaluation or Fitness Function}
In our case, the fitness function for an agent is defined as follows:

\[
f(\text{agent}) = \text{CR} \cdot \text{CC} + \text{DR} \cdot \text{D} - \text{TP} \cdot (\text{MT} - \text{TL})
\]

Where:
\begin{align*}
\text{CR} & : \text{Coin Reward} \\
\text{CC} & : \text{Collected Coins} \\
\text{DR} & : \text{Distance Reward} \\
\text{D} & : \text{Distance Traveled} \\
\text{TP} & : \text{Time Penalty} \\
\text{MT} & : \text{Max Time Allowed} \\
\text{TL} & : \text{Time Left}
\end{align*}

The fitness function employed in the MarioGA methodology serves as a critical component for evaluating the performance of each agent in playing the Super Mario Bros game. It calculates an agent's fitness score based on several key factors. Firstly, it rewards the agent for collecting coins, where the reward for each collected coin is denoted as CR (Coin Reward), and the actual number of coins collected by the agent during gameplay is represented as CC (Collected Coins). Secondly, it encourages the agent to travel a greater distance within the game, with DR (Distance Reward) as the reward factor and D (Distance Traveled) as the agent's recorded distance. Lastly, the fitness function introduces a penalty term to discourage agents from consuming excessive time during gameplay, where TP (Time Penalty) represents the penalty factor, MT (Max Time Allowed) denotes the maximum time allowed for completing the game level, and TL (Time Left) indicates the remaining time available to the agent. The fitness function balances these factors to assess an agent's overall performance, with the aim of evolving agents over generations to maximize their fitness scores and game-playing capabilities.

\subsection{Selection Criteria}

\begin{figure}
  \centering
  \begin{tikzpicture}[
    node distance=2cm,
    block/.style={rectangle, draw, text width=4cm, text centered, rounded corners, minimum height=1.5cm},
    line/.style={draw, -latex'},
  ]

    \node [block] (init) {Initialize Population};
    \node [block, below of=init] (chooseT) {Choose Tournament Size (\(T\))};
    \node [block, below of=chooseT] (select) {Select Candidates for Tournament};
    \node [block, below of=select] (evaluate) {Evaluate Fitness};
    \node [block, below of=evaluate] (winner) {Choose Winner};
    \node [block, below of=winner] (nextgen) {Select Winner as Parent for Next Generation};
    \node [block, below of=nextgen] (repeat) {Repeat for Desired Parents};

    \path [line] (init) -- (chooseT);
    \path [line] (chooseT) -- (select);
    \path [line] (select) -- (evaluate);
    \path [line] (evaluate) -- (winner);
    \path [line] (winner) -- (nextgen);
    \path [line] (nextgen) -- (repeat);

  \end{tikzpicture}
  \caption{Tournament Selection Process}
  \label{fig:ts}
\end{figure}
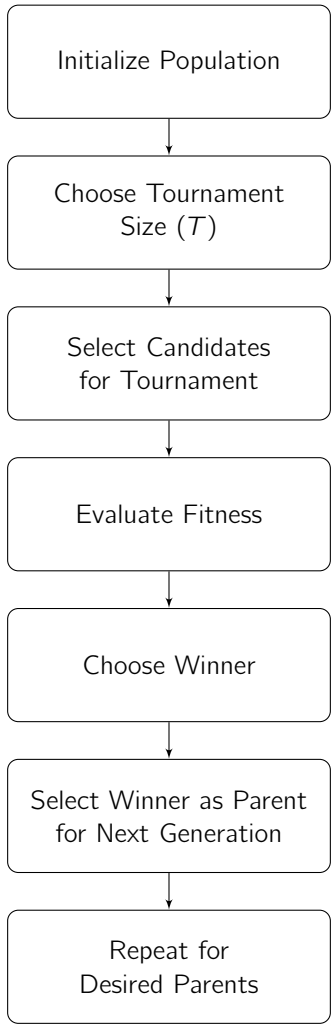

Our methodology for MarioGA uses a tournament selection mechanism as part of its genetic algorithm. In the $play_generation$ function, the code selects the elite agent (the one with the highest fitness) as a parent without making any changes to its moves. Then, for the rest of the population, it creates new agents through crossover and mutation. This approach is often referred to as "tournament selection" because it pits individuals against each other in small tournaments to determine which individuals will be selected as parents for the next generation. The tournament selection process helps maintain diversity within the population while favoring individuals with higher fitness.
To implement tournament selection (as displayed in Figure \ref{fig:ts}) in a MarioGA, we follow the algorithmic description below:

\begin{enumerate}
  \item Initialize a population \(P\) with individuals.
  \item Choose a tournament size \(T\) (the number of individuals competing in each tournament).
  \item Repeat the following steps for each parent you need to select:
  \begin{enumerate}
    \item Randomly select \(T\) individuals (candidates) from the current population \(P\).
    \item Calculate the fitness of each candidate.
    \item Choose the candidate with the highest fitness as the winner.
    \item The winner is selected as a parent for the next generation.
  \end{enumerate}
  \item Repeat step 3 for the desired number of parents.
\end{enumerate}

\subsection{Recombination or Crossover}
\begin{figure}[htp]
  \centering
  \includegraphics[width=0.99\columnwidth]{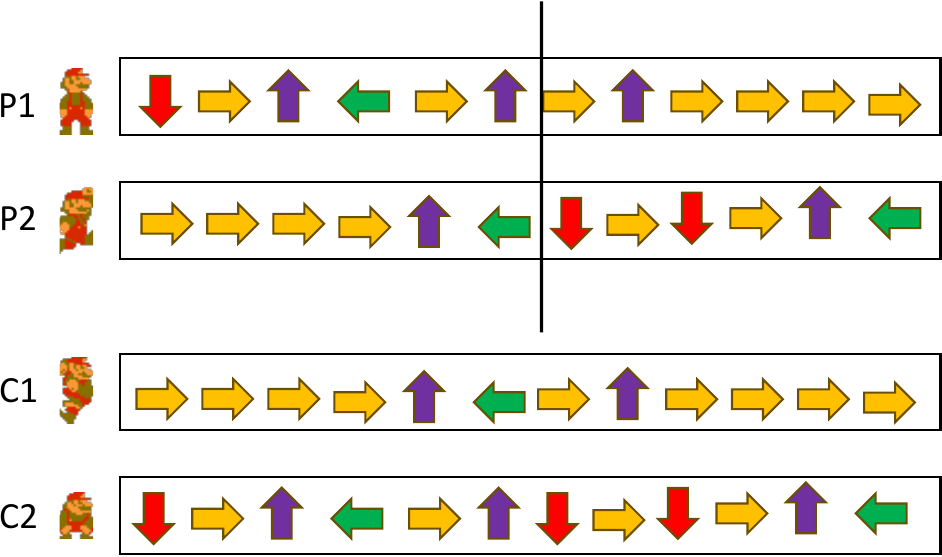}
  \caption{Single point crossover at cross-over point at 50\% of the total size of an Individual sequence of moves.
  \label{fig:crossover}
  }
\end{figure}

In the context of MarioGA, the crossover (recombination) operator is used to create new agent individuals (offspring) by merging genetic information from two parent agents. Each agent is represented by a sequence of moves that determine its gameplay strategy. The crossover operator mimics genetic recombination in natural reproduction, allowing promising strategies from different agents to be combined to potentially produce more successful offspring. Example shown in Figure \ref{fig:crossover} displays how it is working in MarioGA.

The crossover type used is One-Point Crossover in MarioGA which is described as:

\begin{itemize}
  \item Select two parent agents, each represented by a sequence of moves.
  \item Choose a crossover point within the move sequences.
  \item Exchange move sequences to create two new agent offspring:
    \begin{itemize}
      \item Offspring 1: Combines moves from Parent 1 before the crossover point with moves from Parent 2 after the crossover point.
      \item Offspring 2: Combines moves from Parent 2 before the crossover point with moves from Parent 1 after the crossover point.
    \end{itemize}
  \item One-point crossover in this context enables the genetic algorithm to explore different gameplay strategies by mixing and matching moves from successful parent agents.
\end{itemize}

\subsection{Mutation}
Mutation is a fundamental operator in genetic algorithms (GAs) that introduces randomness and diversity into the population of individuals (genomes). It plays a critical role in maintaining genetic diversity within the population, preventing premature convergence to sub-optimal solutions, and enabling the exploration of new areas of the search space. In MarioGA, mutation is implemented as follows:

1) Mutation Rate ($mutation\_rate$): A mutation rate is defined at the beginning, indicating the probability of a move being mutated. The mutation rate is set to 0.8 (80\%) for our case, which means there is an 80\% chance that a move will be mutated. 2) $mutate\_moves$ Function: The $mutate\_moves$ function is responsible for introducing mutations into an agent's moves. It selects a starting index in the list of moves based on the percentage of moves that are mutable ($moves\_mutable$). Then, for each move from that index to the end of the list, it randomly decides whether to replace the move with a random action sampled from the environment's action space, based on the mutation rate. 3) Mutation of Moves: Mutation occurs at the level of individual moves within an agent's sequence of actions. The algorithm iterates through the list of moves, and for each move, it checks if a random number is less than the mutation rate. If it is, the move is replaced with a random action. 

Hence, mutation is performed by randomly replacing a subset of an agent's moves with new, randomly selected actions from the environment's action space, and this mutation is a type of random point mutation.

\begin{figure}[htp]
  \centering
  \includegraphics[width=0.99\columnwidth]{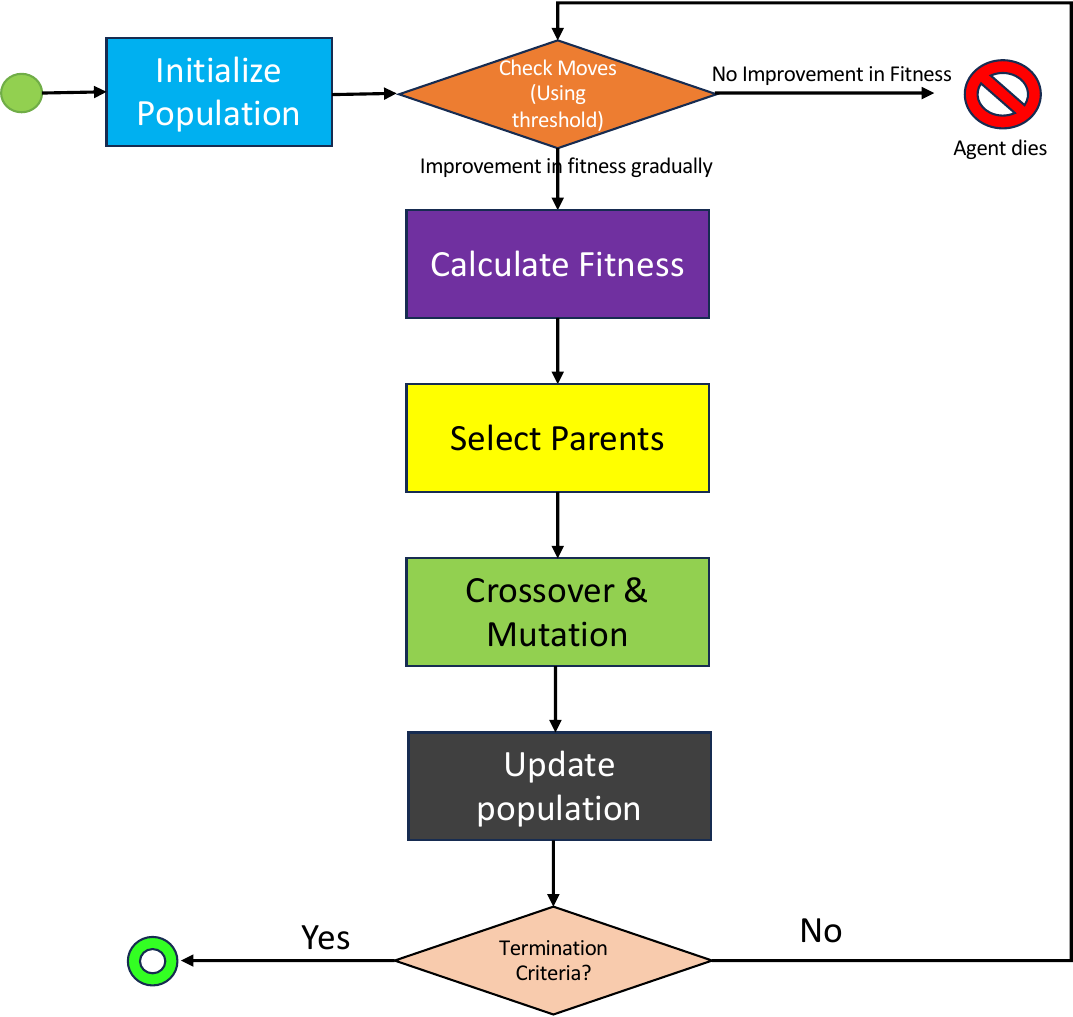}
  \caption{Full working pipeline of the proposed methodology for MarioGA.
  \label{fig:marioga_pipeline}
  }
\end{figure}

\subsection{Termination and Working Algorithm}
The full working pipeline of proposed methodology is shown in Figure \ref{fig:marioga_pipeline}. Additionally, the working algorithm is present in Algorithm \ref{algo:algo1}. The termination criteria are based on the following conditions:

1) No Improvement in Fitness: If an agent fails to show any improvement in its fitness over a certain number of moves (specified by $moves\_to\_check$), it is terminated. This condition ensures that agents evolve efficiently and do not waste computational resources on stagnant individuals.

2) Generation Limit: The code runs for a predefined number of generations ($generation\_amount$) to control the overall duration of the evolutionary process. Once this limit is reached, the evolution process terminates, and the best-performing agent is selected as the final result.

The combination of these criteria helps to ensure that the genetic algorithm converges towards improved solutions and avoids unnecessary computation when it reaches a point where further evolution is unlikely to yield better results.

\begin{algorithm}
    \KwData{Initial population of Mario agents}
    \KwResult{Evolved population of Mario agents}
    \SetKwInOut{Input}{Input}
    \SetKwInOut{Output}{Output}
    \Input{Maximum generations, fitness threshold, mutation rate}
    \Output{Optimal Mario agent}
    \BlankLine
    
    \BlankLine
    \textbf{BEGIN}
    
    \BlankLine
    \textbf{INITIALIZE} population with random Mario agents\;
    \textbf{EVALUATE} fitness for each Mario agent\;
    
    \Repeat{TERMINATION CONDITION is satisfied}{
        \textbf{SELECT} parents based on fitness\;
        \textbf{RECOMBINE} pairs of parents to create offspring\;
        \textbf{MUTATE} the moves of resulting offspring\;
        \textbf{EVALUATE} fitness for the new generation of Mario agents\;
        \textbf{SELECT} individuals for the next generation\;
    }
    
    \textbf{END}
    \caption{Mario Genetic Algorithm (MarioGA)}
      \label{algo:algo1}
\end{algorithm}

\section{Proposed Methodology of MarioNE}
MarioNE leverages NE to evolve increasingly proficient Mario agents over generations. This section explores the key components and definitions of MarioNE.

\subsection{Neuroevolution (NE)}
Neuroevolution is a subfield of artificial intelligence and machine learning that applies evolutionary algorithms to optimize artificial neural networks (ANNs) for solving complex tasks. In the context of MarioNE, NE is used to evolve neural network-based Mario agents.

\begin{figure*}[htp]
  \centering
  \includegraphics[width=1.99\columnwidth]{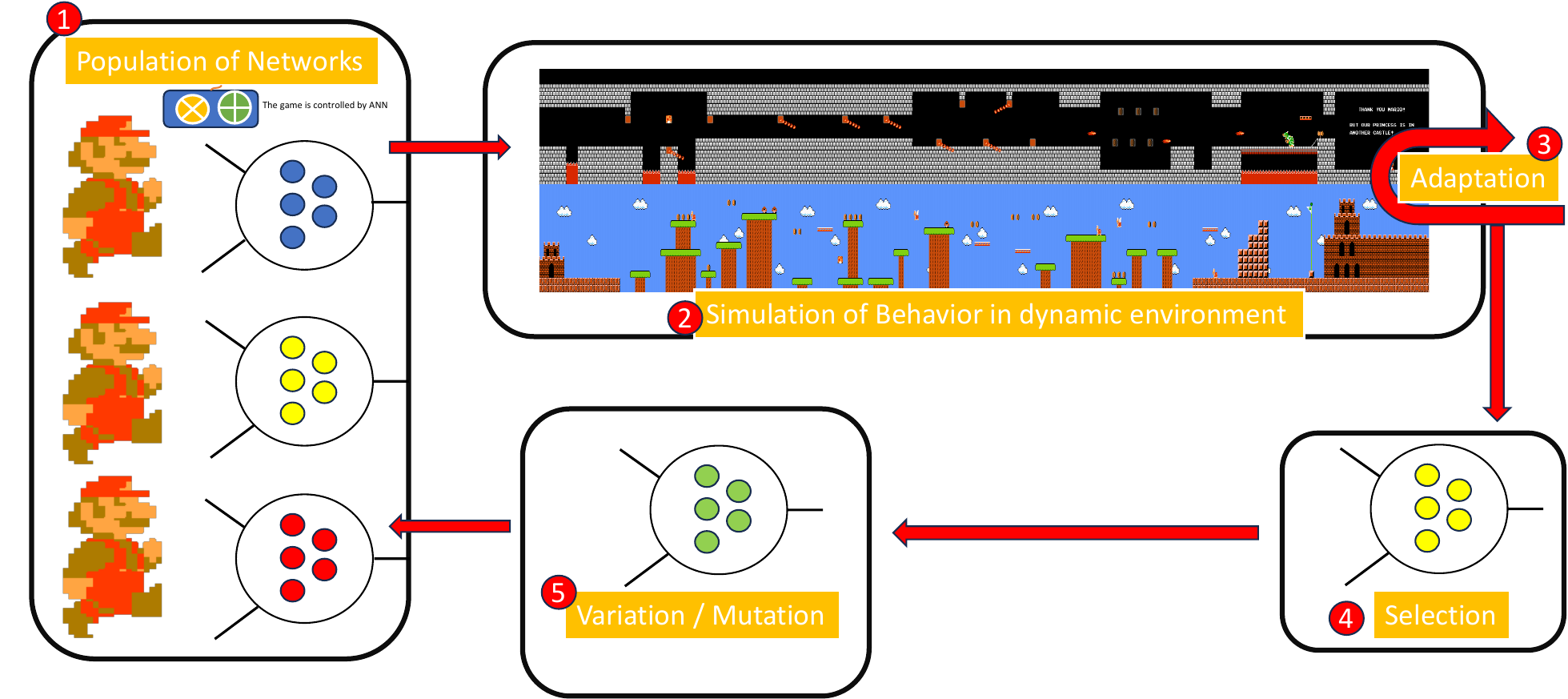}
  \caption{Full working pipeline of the proposed methodology for MarioNE.
  \label{fig:ne}
  }
\end{figure*}
\subsection{Neuro Evolution Operators}
1) Population: The population consists of a group of Mario agents, each represented by a unique neural network. During NE, multiple agents coexist within the population, and their performance is evaluated to select the most fit individuals. 2) Fitness Function: The fitness function quantifies the performance of a Mario agent in the game. It assigns a numerical score based on how well the agent accomplishes specific objectives, such as collecting coins, avoiding obstacles, and reaching the level's end. 3) Selection: The selection process in NE involves choosing the fittest Mario agents from the population to become parents for the next generation. Agents with higher fitness scores have a greater chance of being selected. 4) Crossover: Crossover is a genetic operator where pairs of parent Mario agents exchange genetic material (neural network parameters) to create offspring. It introduces diversity in the neural networks of the next generation. 5) Mutation: Mutation is another genetic operator that introduces small random changes to the neural networks of Mario agents. It helps explore new behaviors and prevents the population from converging prematurely. 6) Termination Criteria: Termination criteria are conditions that determine when the NE process should stop. Common termination criteria include reaching a predefined fitness threshold, a maximum number of generations, or a fixed computational budget. 7) Evolutionary Progress: The evolutionary progress in MarioNE is measured by observing how the fitness of the population changes across generations. As the NE process continues, Mario agents evolve to become more proficient at playing the game.

\subsection{MarioNE Methodology}

In this section, we detail the methodology for employing Neuroevolution (NE) to evolve intelligent agents for playing Super Mario Bros. MarioNE aims to optimize neural network-based agents to achieve better performance in the game while adhering to the fitness function introduced in MarioGA. The following steps outline the MarioNE methodology:

\subsection{Neuroevolution Framework}

To implement MarioNE, we utilize a neuroevolution framework that provides tools for evolving neural network-based agents. We choose the NEAT (NeuroEvolution of Augmenting Topologies) algorithm due to its ability to evolve both neural network weights and structures. NEAT maintains a population of neural networks with varying complexities, allowing for the exploration of different strategies.

\subsection{Neural Network Architecture}

We design a neural network architecture that serves as the brain of our Mario agents. The neural network takes game state information as input and produces actions as output. The architecture includes layers for processing input data, hidden layers for learning complex patterns, and output layers for mapping to Mario's actions. We employ activation functions like the rectified linear unit (ReLU) to introduce non-linearity.

\subsection{Fitness Function}

The fitness function in MarioNE mirrors the one used in MarioGA to align with the objectives of collecting as many coins as possible, traveling a large distance, and minimizing time spent in the game. The fitness function is defined as follows:

\begin{equation}
\label{equation1}
    \text{Fitness} = \alpha \cdot \text{Coins} + \beta \cdot \text{Distance} - \gamma \cdot \text{Time}
\end{equation}

Where:
\begin{itemize}
    \item $\alpha$, $\beta$, and $\gamma$ are trade-off parameters.
    \item Coins represent the number of collected coins.
    \item Distance is the progress made through the level.
    \item Time is the time taken to complete the level, with a penalty.
\end{itemize}

The trade-off parameters allow us to balance the importance of each objective in the fitness function.
\
\subsection{Neuroevolution Process}

The core of MarioNE is the neuroevolution process, which optimizes Mario agents' neural networks over generations. We employ the following steps:

\begin{enumerate}
    \item {Initialization}: Initialize a population of Mario agents with randomly generated neural networks.
    \item {Evaluation}: Evaluate the fitness of each agent by running them through Super Mario Bros.
    \item {Selection}: Select agents for reproduction based on their fitness scores. Higher fitness leads to a higher chance of selection.
    \item {Crossover}: Create offspring by combining the neural networks of selected parents. Our implementation (discussed in later sections of the paper) also facilitates crossover with the preservation of innovation history.
    \item {Mutation}: Introduce small random changes to the offspring's neural networks to maintain diversity.
    \item {Evaluation (Next Generation)}: Evaluate the fitness of the new generation of agents in the game environment.
    \item {Termination}: Repeat the process for multiple generations or until a termination criterion (e.g., No improvement in fitness) is met.
\end{enumerate}

Through this iterative process, Mario agents evolve to exhibit increasingly intelligent gameplay behaviors. A simple pipeline of the working of NeuroEvolution is shown in Figure \ref{fig:ne}.


\section{Implementation details and Experimental Setup of MarioGA}




We implement MarioGA using python and supporting libraries of Gym-OpenAI. Following are critical details of our implementation.

\subsection{Agent Class}

The code defines an \texttt{Agent} class representing an individual agent. Each agent possesses the following attributes:

\begin{itemize}
    \item \texttt{fitness}: A measure of an agent's performance.
    \item \texttt{moves}: A sequence of actions representing the agent's gameplay.
    \item \texttt{moves\_used}: The count of moves executed by the agent.
\end{itemize}


\begin{table*}[h]
\caption{Genetic Operators in MarioGA}
\label{tab:genetic-operators}
\centering
\begin{tabular}{|p{2.8cm}|p{10cm}|p{3cm}|}
\hline
\textbf{Genetic Operator} & \textbf{Operator Description} & \textbf{Operator Value} \\
\hline
Representation & List or Sequence of actions and moves & ['right', 'A', 'B', 'left','down','up'] \\
\hline
Crossover & One-Point Crossover & 50\% \\
\hline
Mutation & Flip and Change (adaptive mutation with a fraction for fittest) & 0.01±0.005\\
\hline
Population & Collection of all individuals & 20 \\
\hline
Generation Size & Number of iterations to evolve agents & 1000 \\
\hline
Moves Amount & Number of all moved that each agent posses before gameplay & 5000 \\
\hline
Moves to Check & Number of moves that do not lead to an increase in fitness, (agent stuck in suboptimum) & 30 \\
\hline
Moves Mutable & This value determines how much percentage of moves are going to be mutated or perturbed & 80\% \\
\hline
Coin Reward (+) & This gets multiplied with the amount of coins collected by Mario in gameplay & 10 \\
\hline
Distance Reward (+) & This gets multiplied with the distance traveled by Mario in gameplay & 0.1 \\
\hline
Time Penalty (-) & This gets multiplied with the total time quantum in gameplay & 0.8 \\
\hline
Elitism & We preserve at least some best agent from the last generation & 1 \\
\hline
Number of Offspring & Children produced after crossover or recombination operator & 2 \\
\hline
Initialisation & How are we initializing the agents or agent sequence of moves & Random \\
\hline
Termination & We use two types of termination criteria & 'generation size', 'moves to check' \\
\hline
\end{tabular}
\end{table*}

\subsection{Genetic Algorithm Components}

\begin{itemize}
    \item \texttt{create\_population()}: This function generates a population of agents. It creates agents with random moves.
    
    \item \texttt{check\_fitness()}: This function calculates an agent's fitness based on various factors, including coins collected, distance traveled, and time survived. It also implements a mechanism for terminating agents with no fitness progress.
    
    \item \texttt{mutate\_moves()}: Agents undergo random mutations in their move sequences. The mutation rate and the portion of moves that can be mutated are configurable.
    
    \item \texttt{play\_generation()}: This function simulates a generation of agents' gameplay. It selects the fittest agent, preserves it, and applies crossover and mutation to the rest of the population. The mutation rate is adjusted based on fitness progress.
    
    \item \texttt{record\_player()}: It records the gameplay of an agent, stores images, and calculates its fitness. The best agent's gameplay is captured for animation.
\end{itemize}

\subsection{Main Loop}

The main loop iterates through generations and evolves the population of agents using the genetic algorithm. The key steps include:

\begin{enumerate}
    \item Creating a new population of agents.
    
    \item Simulating the generation by playing agents, calculating fitness, and evolving the population.
    
    \item Saving the best agent's gameplay as animations in a dedicated folder.
    
    \item Adjusting mutation rates based on fitness progress.
    
    \item Continuously monitoring and recording the highest fitness achieved in each generation.
\end{enumerate}

\subsection{Parameters}

The code allows various parameters to be configured, such as mutation rate, population size, and fitness components (coin reward, distance reward, time penalty). These parameters enable fine-tuning of the genetic algorithm's behavior.

\subsection{Custom Starting Agent}

The \texttt{custom\_starting\_agent()} function enables the use of a predefined agent to start from a specific point in the training process.

Overall, our Python code implements a comprehensive genetic algorithm (MarioGA) for training an agent to play SMB. It includes various functions for agent management, fitness calculation, and population evolution. The code allows for extensive customization and records the best agent's gameplay as animations for analysis. For any code related or explanation of the variables, we refer code. Moreover, we present a tabular summary in Table \ref{tab:genetic-operators} of the implementation of MarioGA with description of the GA. 

\section{Implementation details and Experimental Setup of MarioNE}

\begin{table*}[h]
\caption{Genetic Operators in MarioNE}
\label{tab:genetic-operators-marioNE}
\centering
\begin{tabular}{|p{2.8cm}|p{8cm}|p{5cm}|}
\hline
\textbf{Genetic Operator} & \textbf{Operator Description} & \textbf{Operator Value} \\
\hline
Representation & Neural Network Architecture & Dynamic, Evolving \\
\hline
Initialization & Random Initialization of Neural Networks & Random Weights \\
\hline
Fitness Evaluation & Assessment of Agent Performance & Coin Collection, Distance Traveled, Time Penalty \\
\hline
Population & Collection of Neural Networks & 1 (Configurable) \\
\hline
Generations & Number of Iterations in Evolution & 1000 (Configurable) \\
\hline
Mutation Rate & Probability of Weight Mutation & 10\% (Configurable) \\
\hline
Number of Actions & Available Actions for Mario & Variable (e.g., RIGHT\_ONLY) \\
\hline
Crossover & Reproduction Through Genetic Recombination & One-Point Crossover \\
\hline
Selection & Choosing Top Performers for Reproduction & Top 20\% (Configurable) \\
\hline
Termination & Criteria for Ending Evolution & Fitness Threshold, Maximum Generations \\
\hline
\end{tabular}
\end{table*}

The Mario Neuro Evolution (MarioNE) framework leverages NeuroEvolution of Augmenting Topologies (NEAT) to train Mario agents within the context of Super Mario Bros. NEAT is a powerful neuroevolutionary algorithm that evolves neural networks with dynamic architectures, making it well-suited for complex tasks like mastering video games. MarioNE's implementation involves several key components and stages which are also summarised in Table \ref{tab:genetic-operators-marioNE} and Table \ref{table:appendix1}, as outlined below:

\subsection{Neural Network Architecture}
The neural networks in MarioNE are fundamental to agent decision-making. Each neural network is instantiated with a random weight matrix, connecting the input layer (representing the game state) to the output layer (determining Mario's actions). NEAT allows dynamic construction of hidden layers, adding nodes and connections as the networks evolve over generations. This adaptability enables the agents to discover complex strategies through learned behaviors.

\subsection{Initialization and Population}
At the beginning of each NEAT-based evolutionary run, MarioNE initializes a population of neural networks. These networks collectively represent the evolving Mario agents. The population size is a configurable parameter, and typically, it comprises multiple neural networks with varying structures.

\subsection{Fitness Evaluation}
To assess the performance of each agent, a fitness function is employed. The fitness function is designed to optimize specific objectives while penalizing undesired behaviors. In the case of MarioNE, the fitness function incentivizes agents to collect coins, traverse longer distances, and complete levels within minimal time. Simultaneously, it penalizes extended gameplay time. The fitness evaluation process simulates Mario's gameplay using the neural network's decisions and records the achieved fitness scores.

\subsection{NEAT Evolutionary Process}
MarioNE employs the NEAT algorithm to evolve the neural networks over a predefined number of generations. During each generation, the following steps are executed:

\begin{enumerate}
    \item {Selection of Top Performers:} The top-performing agents, typically the top 20\% based on fitness scores, are selected for reproduction. This selective pressure ensures that the most promising agents propagate their genetic information to the next generation.
    
    \item {Crossover and Mutation:} To maintain diversity and facilitate exploration, new agents are created through crossover and mutation operations. In the crossover step, two parent networks contribute to the genetic makeup of a child network, allowing for the exchange of advantageous traits. Mutation introduces variability into the child's neural network structure and connection weights. These operations are governed by NEAT-specific mechanisms.
    
    \item {Population Replacement:} The newly created agents, including the top performers from the previous generation, replace the previous population, forming the next generation. This step ensures that genetic diversity is maintained while enabling further evolution.
\end{enumerate}

\subsection{Termination and Best Agent}
MarioNE iterates through generations until a predefined termination condition is met, such as achieving a certain fitness threshold or reaching a specified maximum number of generations. The best-performing agent, based on the highest fitness achieved, is identified and considered the output of the evolutionary process.

MarioNE's implementation details encompass neural network encoding, initialization, fitness evaluation, selection, crossover, mutation, and termination management. By combining NEAT's flexibility with Super Mario Bros.' gameplay, MarioNE endeavors to evolve proficient Mario agents capable of navigating the game environment adeptly.


\section{Results and Discussions}
In this section we will discuss all results and noteworthy discussions for both the approaches. Firstly, results include worst, average, and best fitness plots performed with five-fold cross validation for each level. In simpler words, we provide fitness plot for both the algorithms that shows generation vs fitness of the MarioGA and MarioNE respectively for world 1 levels 1, 2, 3 and 4. In addition, we also perform same experiment 5 times for each level and average out the results to display in a single plot in order to prove that it was not the lucky case that MarioGA and MarioNE were able to finish levels for at least one experiment. Secondly, we provide additional plots of generation vs the times Mario got stuck in local optimum (cases where there is no progress in fitness, which we discussed in Figure \ref{fig:pipe}). Third, we plot datatset distribution or environment landscape, which in our case is SMB's levels such that it could display the local-optimums where agent could stuck. Fourth, we perform empirical comparisons of both the algorithms on the basis of tweaking various constants ($\alpha$, $\beta$ and $\gamma$) of Fitness Equation \ref{equation1} and plot respective generation vs fitness plots. Fifth, we study the dynamics of the rewards and penalties (in different colors) of different levels within SMB gameplay. This bar plot will provide information of landmarks where it was able to increase the fitness sharply. Additionally, we provide a table where we provide statistics of the gameplay. This statistics include best agent's distance covered, time taken for the agent to finish level, damage taken, kills, coins collected, obstacle's killed or avoided, jumps taken, right move performed, left move performed, life (power ups) discovered and eaten, time taken to find best agent and number of mutations performed in order to reach this best agent. Finally, we discuss running performance of both the algorithms, associating the theoretical time complexity of the algorithms with their experimental performance and interpret all findings. In other words, we compare how the algorithms are expected to perform based on their theoretical analysis (i.e., the time complexity) with how they actually perform in the experiments.

\subsection{Plots for MarioGA}
\begin{figure*}[htp]
  \centering
  \includegraphics[width=1.99\columnwidth]{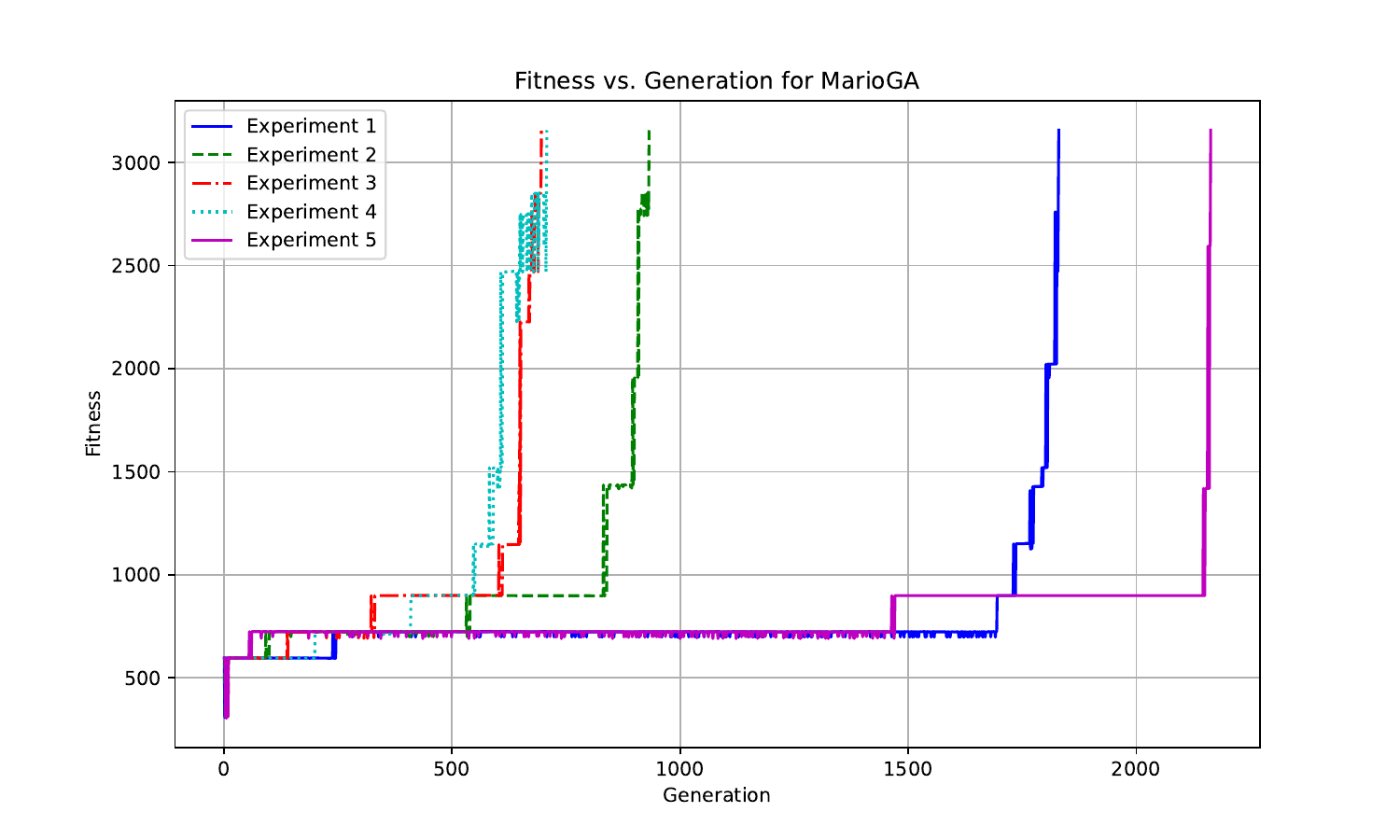}
  \caption{Fitness vs Generation plot for World 1 Level 1. Experiments 3 and 4 were able to converge faster without stuck in some local optimum while experiments 1 and 5 were stuck in local maximum for longer generations. This is where our Mario agent was dying repeatedly and there was no progress for longer generations.
  \label{fig:Algo1plot1World11}
  }
\end{figure*}

\begin{figure*}[htp]
  \centering
  \includegraphics[width=1.99\columnwidth]{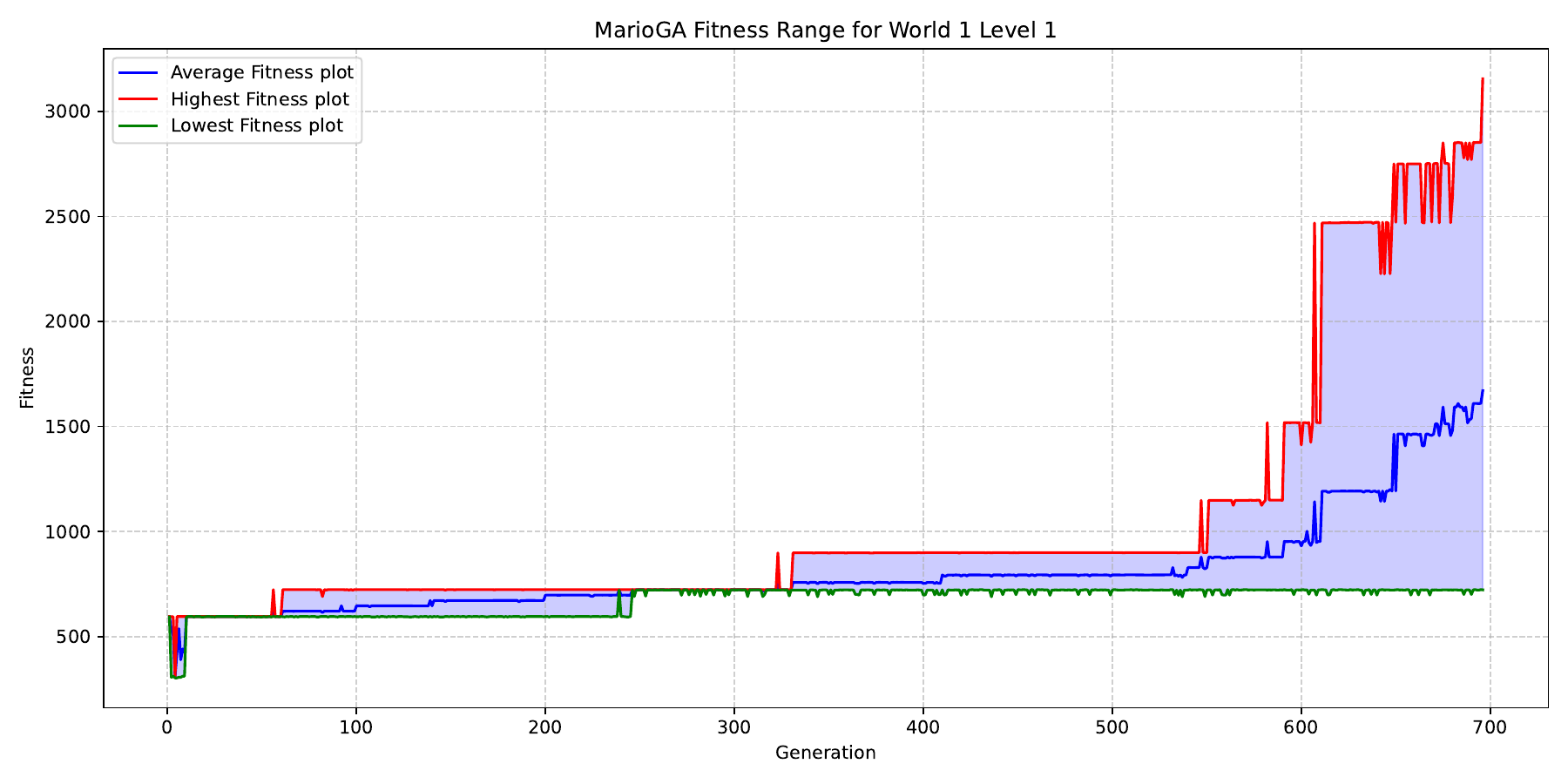}
  \caption{Highest, Average and Lowest fitness compared for 700 generations. This plot says that on average MarioGA was achieving sub-optimum fitness of ±1300 while if we look for highest fitness for 700 generations, it was able to converge with a fitness of ±3000 
  \label{fig:Algo1plot3World11}
  }
\end{figure*}

\begin{figure*}[htp]
  \centering
  \includegraphics[width=1.99\columnwidth]{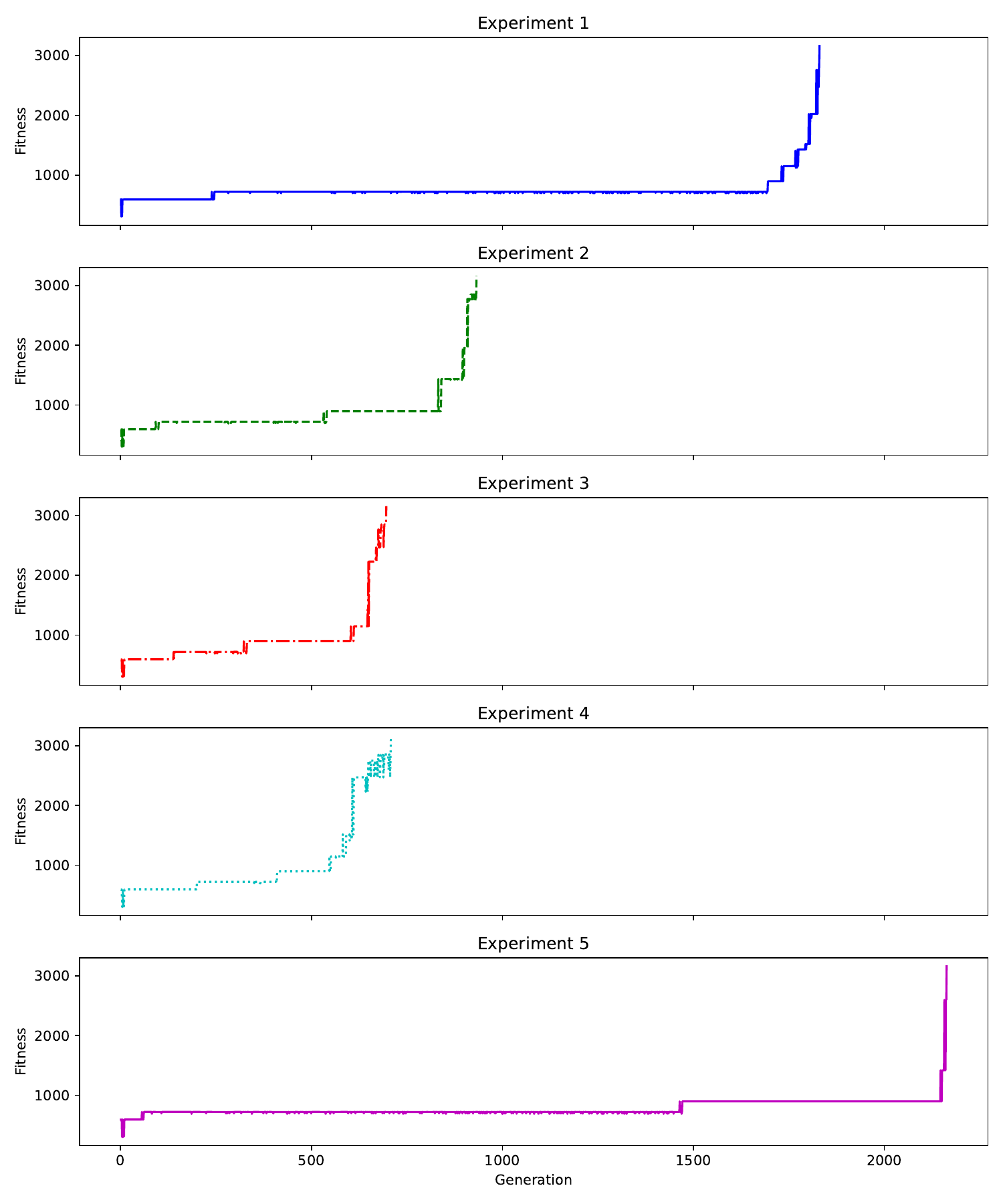}
  \caption{This plot shows five different experiments analysed separately. Overall all experiments were able to finish the level 1.
  \label{fig:Algo1plot2World11}
  }
\end{figure*}

\begin{figure*}[htp]
  \centering
  \includegraphics[width=1.99\columnwidth]{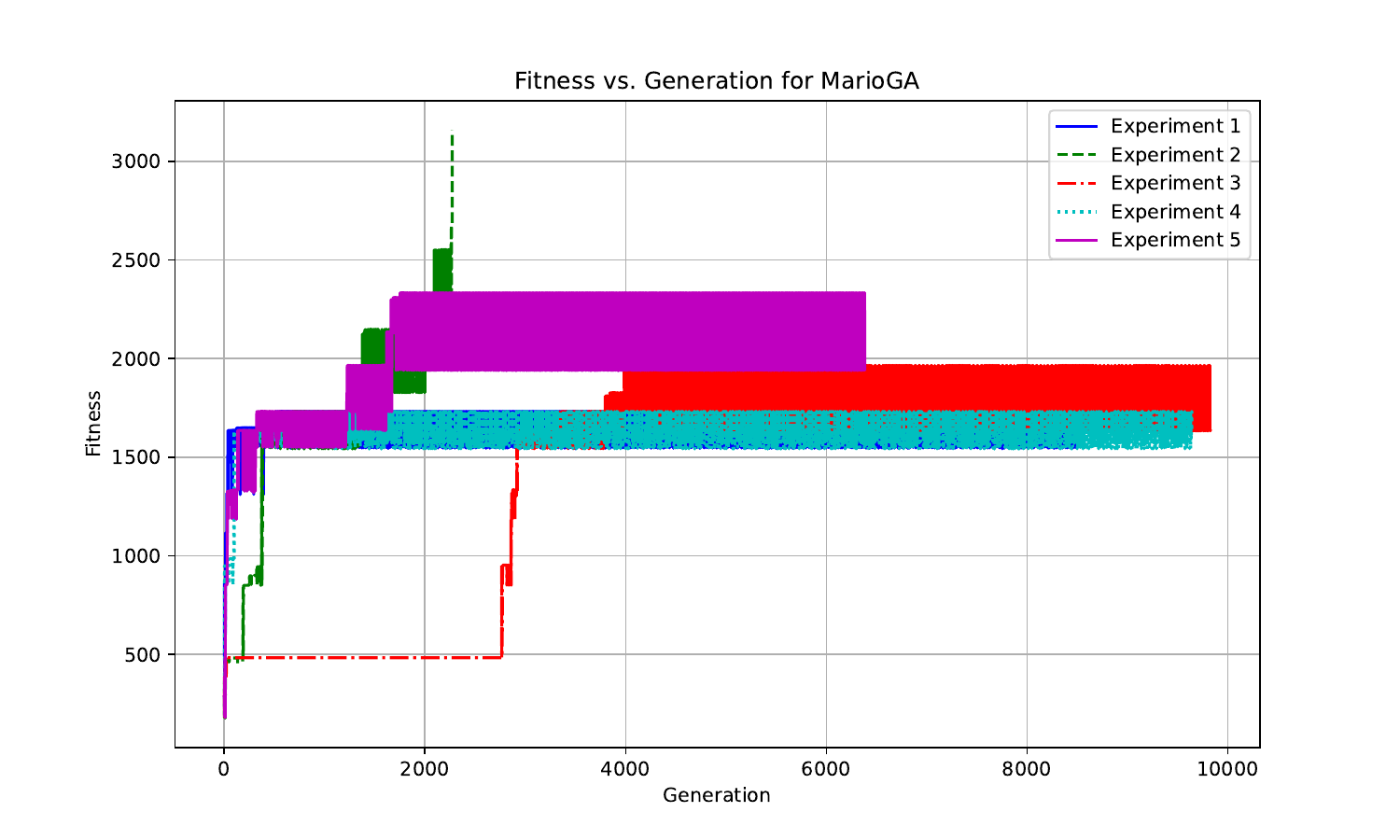}
  \caption{Fitness vs Generation plot for World 1 Level 2. Only Experiment 2 was able to finish the level in very short duration achieving highest fitness of ±3125, however rest of the experiments did not converge even up to 10000 generation size.
  \label{fig:Algo1plot1World12}
  }
\end{figure*}

\begin{figure*}[htp]
  \centering
  \includegraphics[width=1.99\columnwidth]{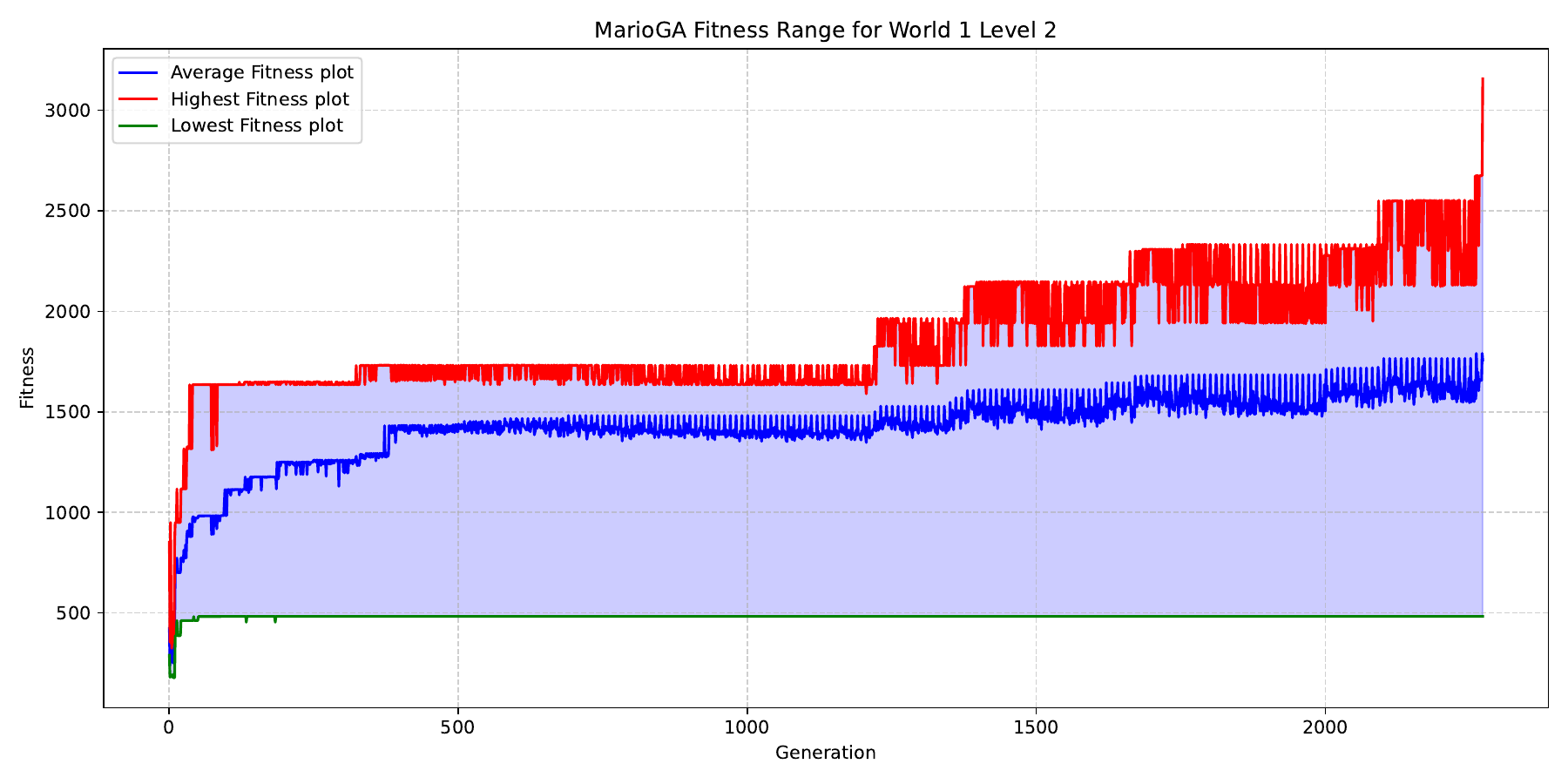}
  \caption{Highest, Average and Lowest fitness compared for 2500 generations. This plot says that on average MarioGA was achieving sub-optimum fitness of ±1700 while if we look for highest fitness for 2500+ generations, it was able to converge with a fitness of ±3125 
  \label{fig:Algo1plot3World12}
  }
\end{figure*}

\begin{figure*}[htp]
  \centering
  \includegraphics[width=1.99\columnwidth]{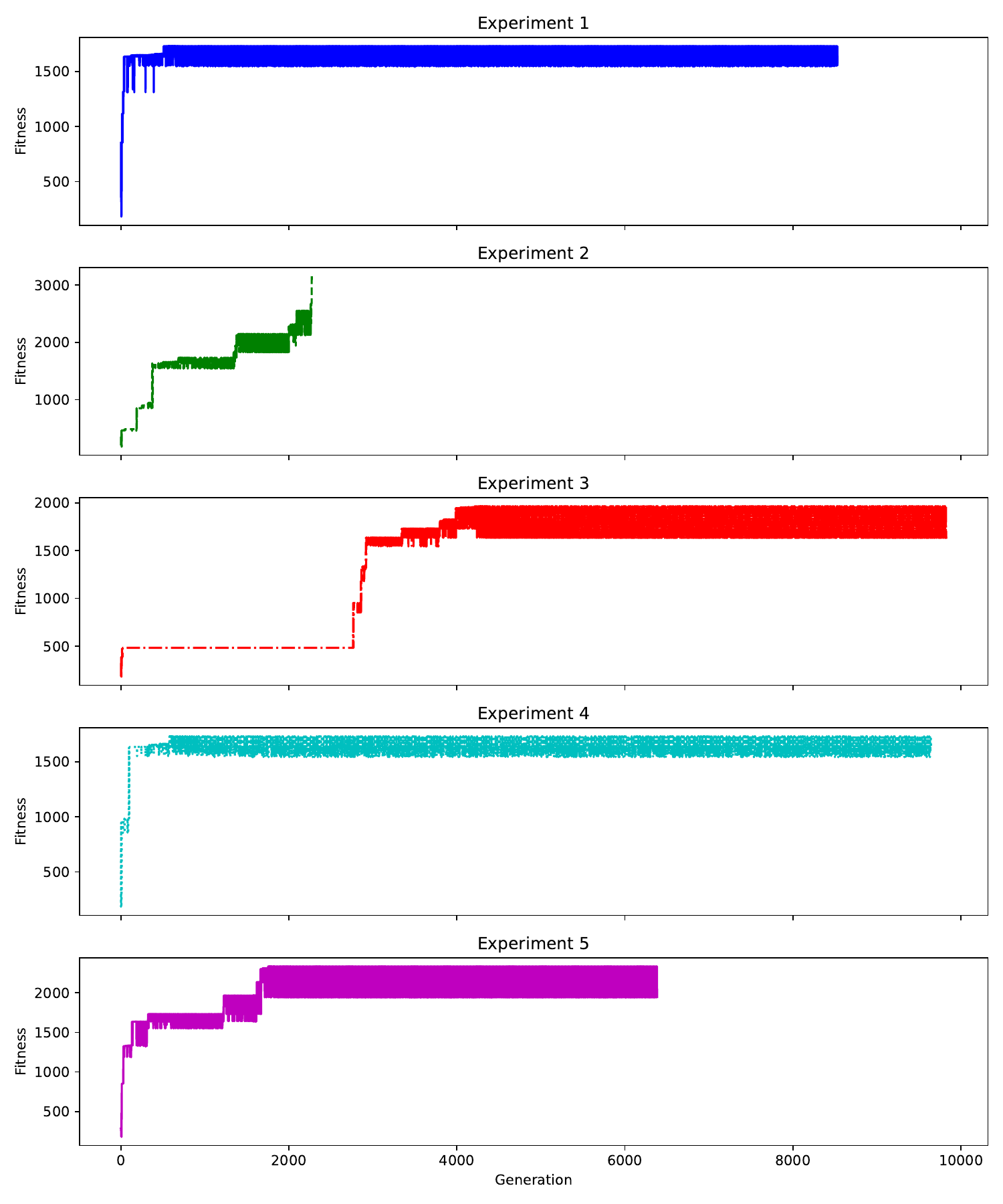}
  \caption{This plot shows five different experiments analysed separately. Overall only experiment 2 was able to finish the level 2.
  \label{fig:Algo1plot2World12}
  }
\end{figure*}

\begin{figure*}[htp]
  \centering
  \includegraphics[width=1.99\columnwidth]{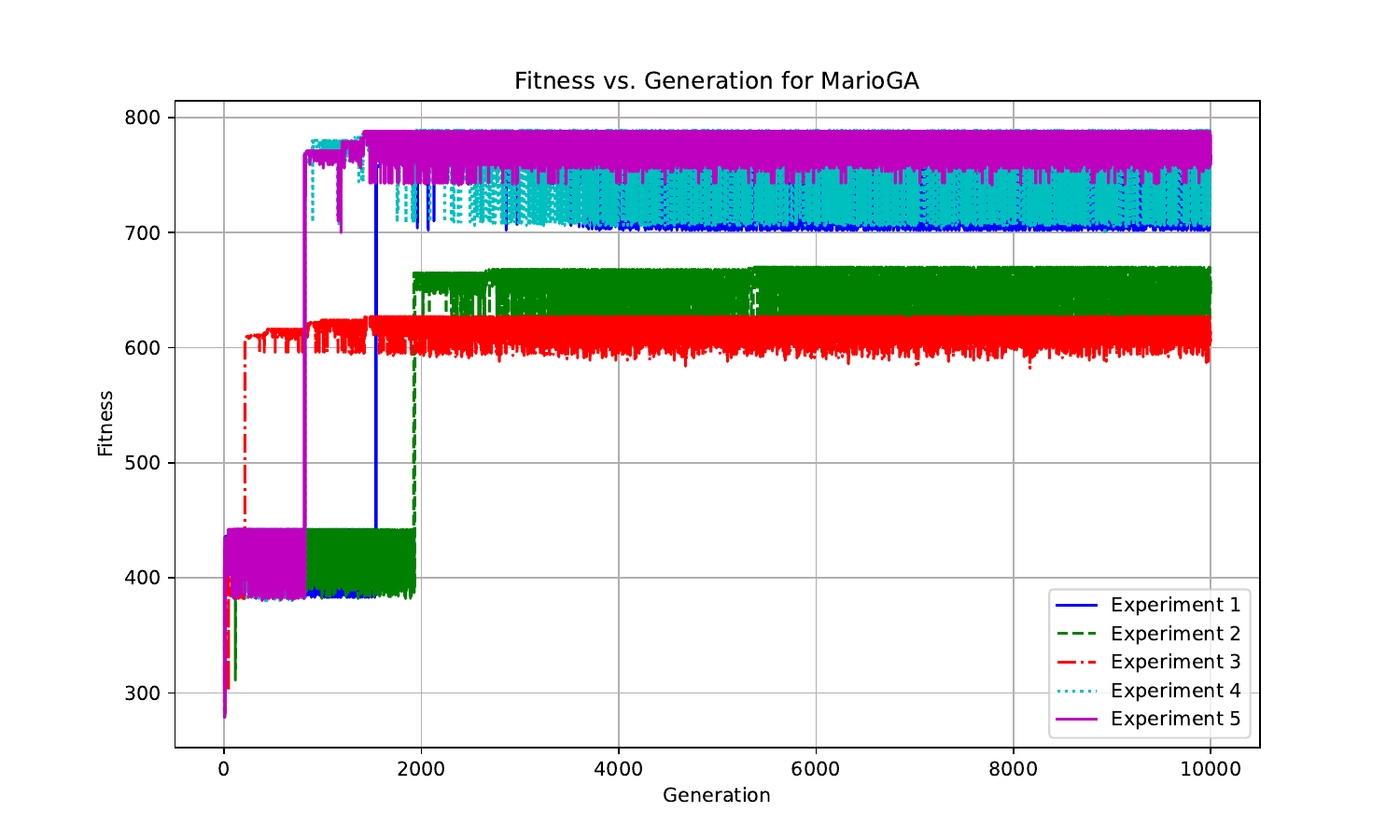}
  \caption{Fitness vs Generation plot for World 1 Level 3. No Experiment was able to finish the level 3, even after running for 10000 generations. 
  \label{fig:Algo1plot1World13}
  }
\end{figure*}

\begin{figure*}[htp]
  \centering
  \includegraphics[width=1.99\columnwidth]{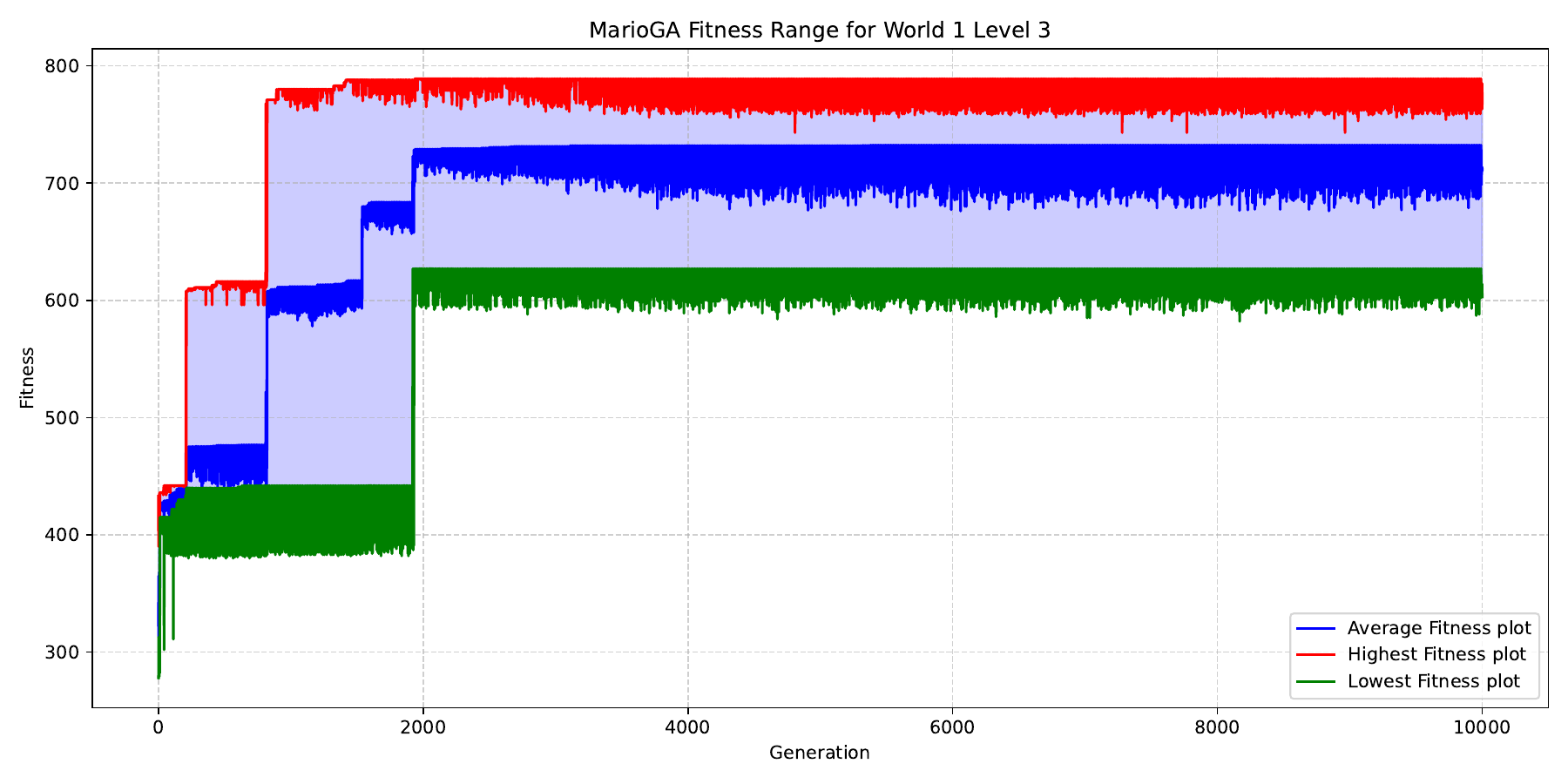}
  \caption{Highest, Average and Lowest fitness compared for 10000 generations. This plot says that on average MarioGA was achieving sub-optimum fitness of ±700 while if we look for highest fitness for 10000 generations, it was not able to converge with a fitness of ±800.
  \label{fig:Algo1plot3World13}
  }
\end{figure*}

\begin{figure*}[htp]
  \centering
  \includegraphics[width=1.99\columnwidth]{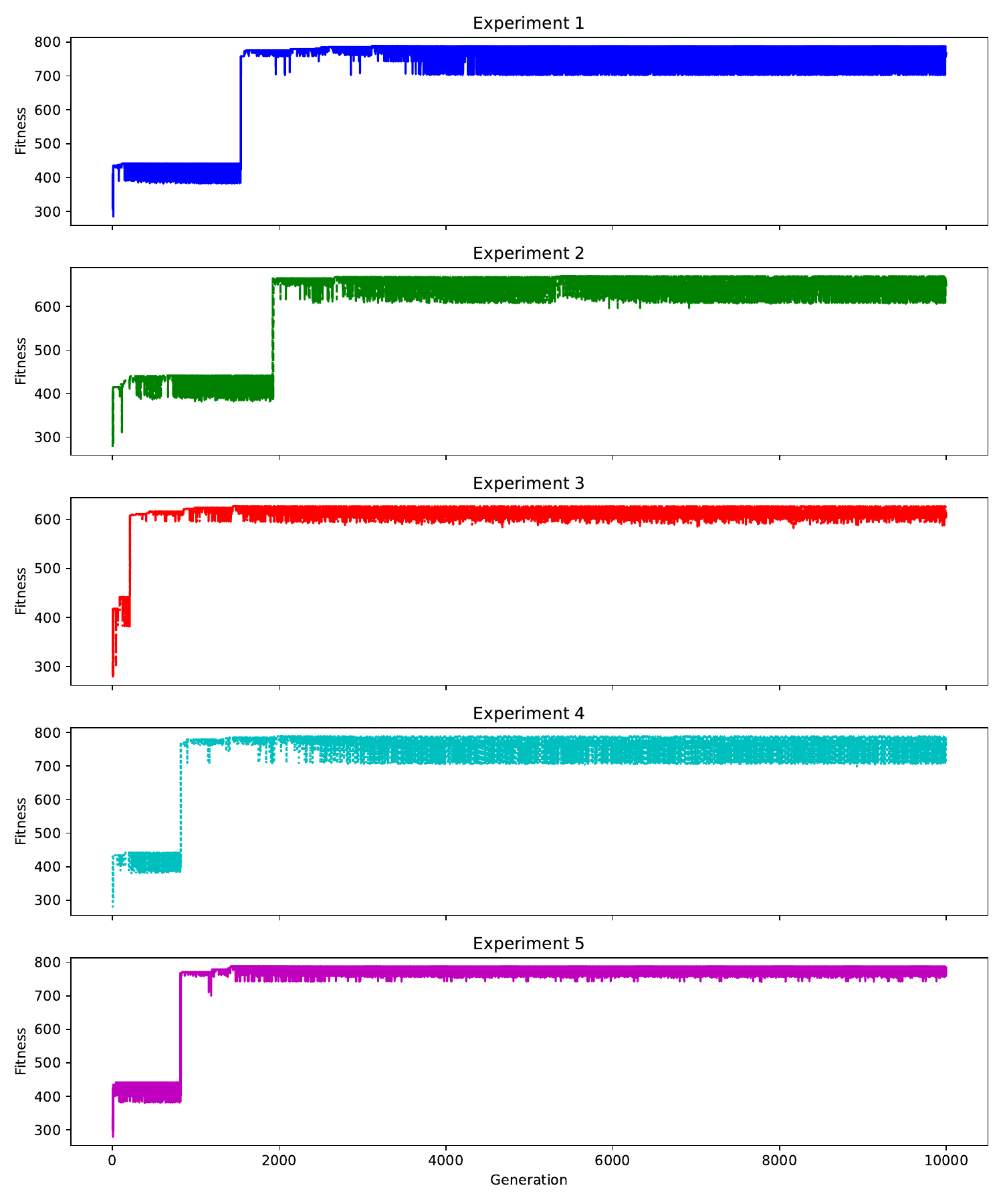}
  \caption{This plot shows five different experiments analysed separately. Overall no experiment was able to finish the level 3.
  \label{fig:Algo1plot2World13}
  }
\end{figure*}

\begin{figure*}[htp]
  \centering
  \includegraphics[width=1.99\columnwidth]{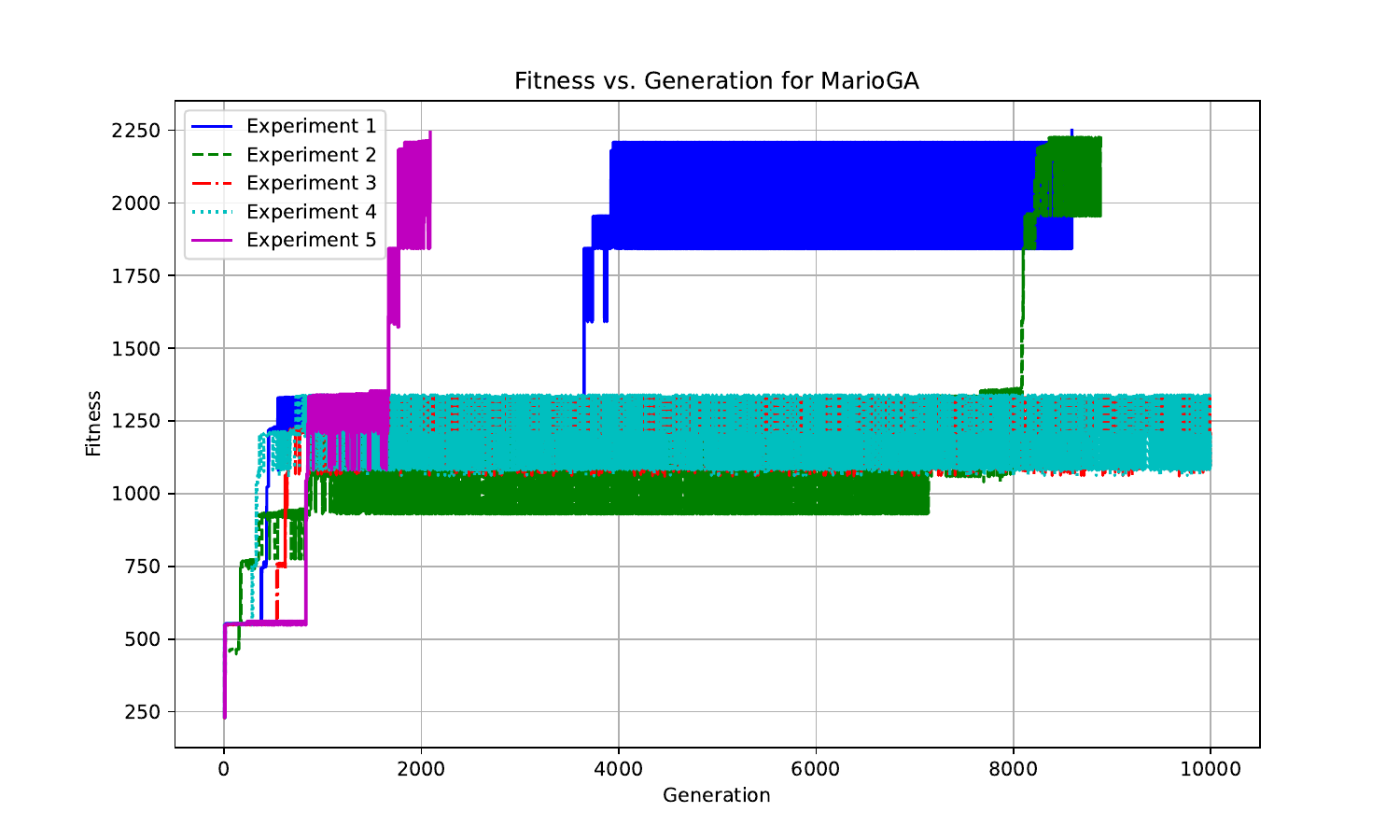}
  \caption{Fitness vs Generation plot for World 1 Level 4. Experiments 1, 2 and 5 were able to finish the gameplay with a fitness of ±2250. Experiment 1 was able to finish gameplay in 8000+ generations, Experiment 2 was able to finish gameplay in ±8200 generations and Experiment 5 was able to finish gameplay in 2200+ generations.
  \label{fig:Algo1plot1World14}
  }
\end{figure*}

\begin{figure*}[htp]
  \centering
  \includegraphics[width=1.99\columnwidth]{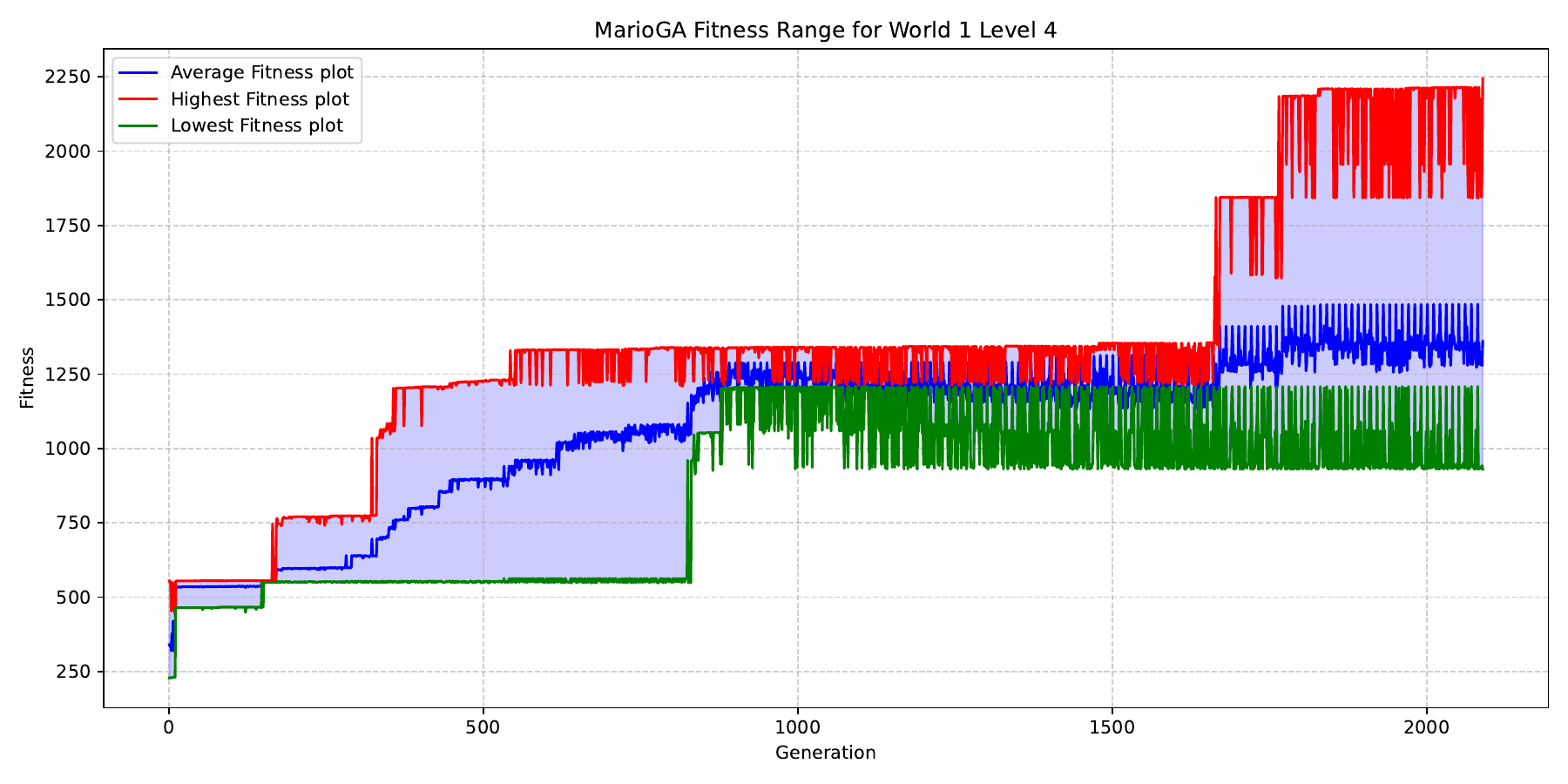}
  \caption{Highest, Average and Lowest fitness compared for 10000 generations. Some of them were able to finish in early generations.
  \label{fig:Algo1plot3World14}
  }
\end{figure*}

\begin{figure*}[htp]
  \centering
  \includegraphics[width=1.99\columnwidth]{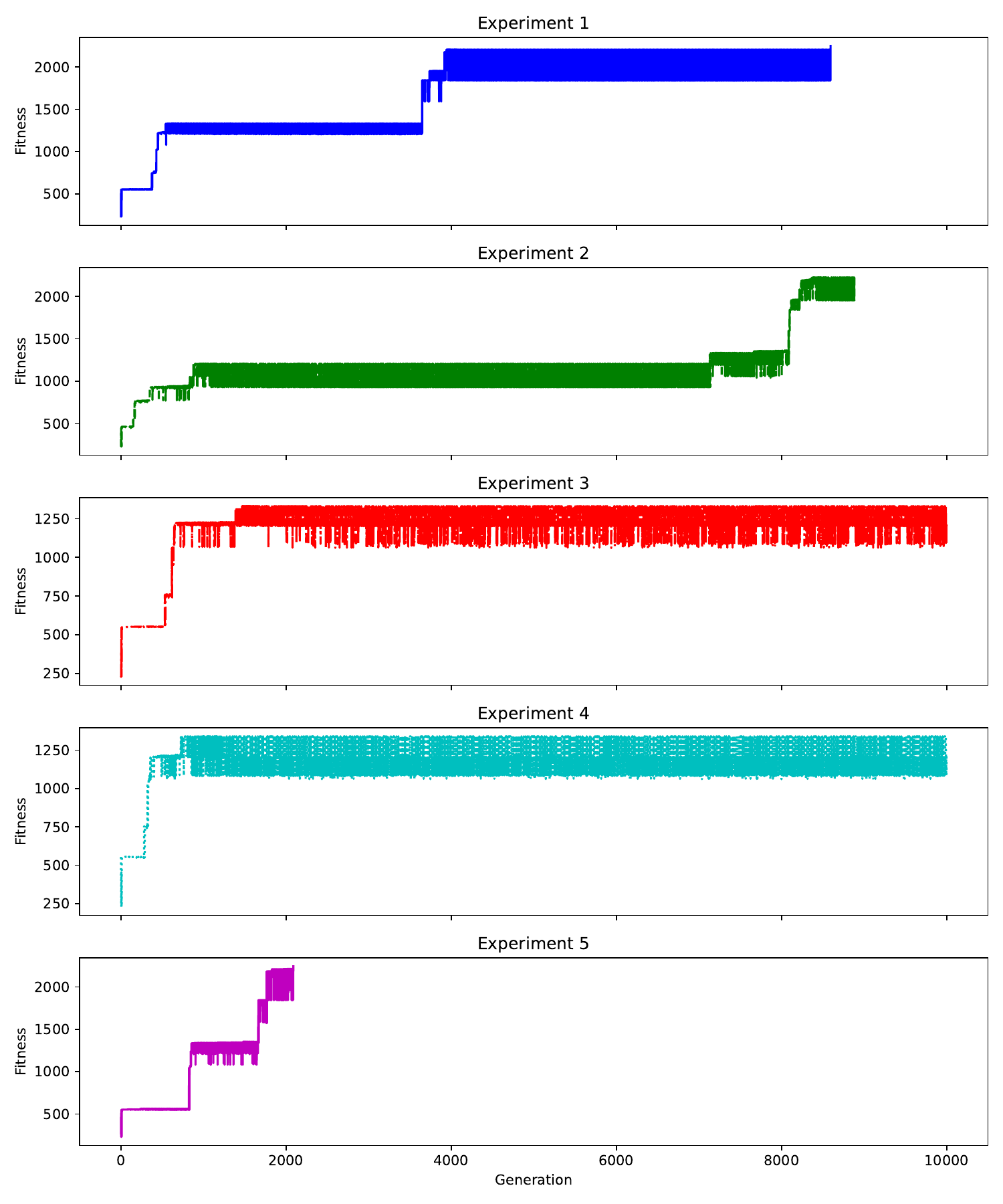}
  \caption{This plot shows five different experiments analysed separately. Overall some experiments were able to finish the level 3 gameplay.
  \label{fig:Algo1plot2World14}
  }
\end{figure*}

In this subsection we particularly focus on plotting Genetic Algorithm experiments. We plot three figures for each world (We perform 5 experiments for each level of world one of the SMB gameplay). For details about plots we refer caption of the plot. For World 1 level 1, Figures \ref{fig:Algo1plot1World11}, \ref{fig:Algo1plot3World11}, \ref{fig:Algo1plot2World11} are shown. For World 1 level 2, Figures \ref{fig:Algo1plot1World12}, \ref{fig:Algo1plot3World12}, \ref{fig:Algo1plot2World12} are shown. For World 1 level 3, Figures \ref{fig:Algo1plot1World13}, \ref{fig:Algo1plot3World13}, \ref{fig:Algo1plot2World13} are shown. For World 1 level 4, Figures \ref{fig:Algo1plot1World14}, \ref{fig:Algo1plot3World14}, \ref{fig:Algo1plot2World14} are shown.

\subsection{Plots for MarioNE}
\begin{figure*}[htp]
  \centering
  \includegraphics[width=1.99\columnwidth]{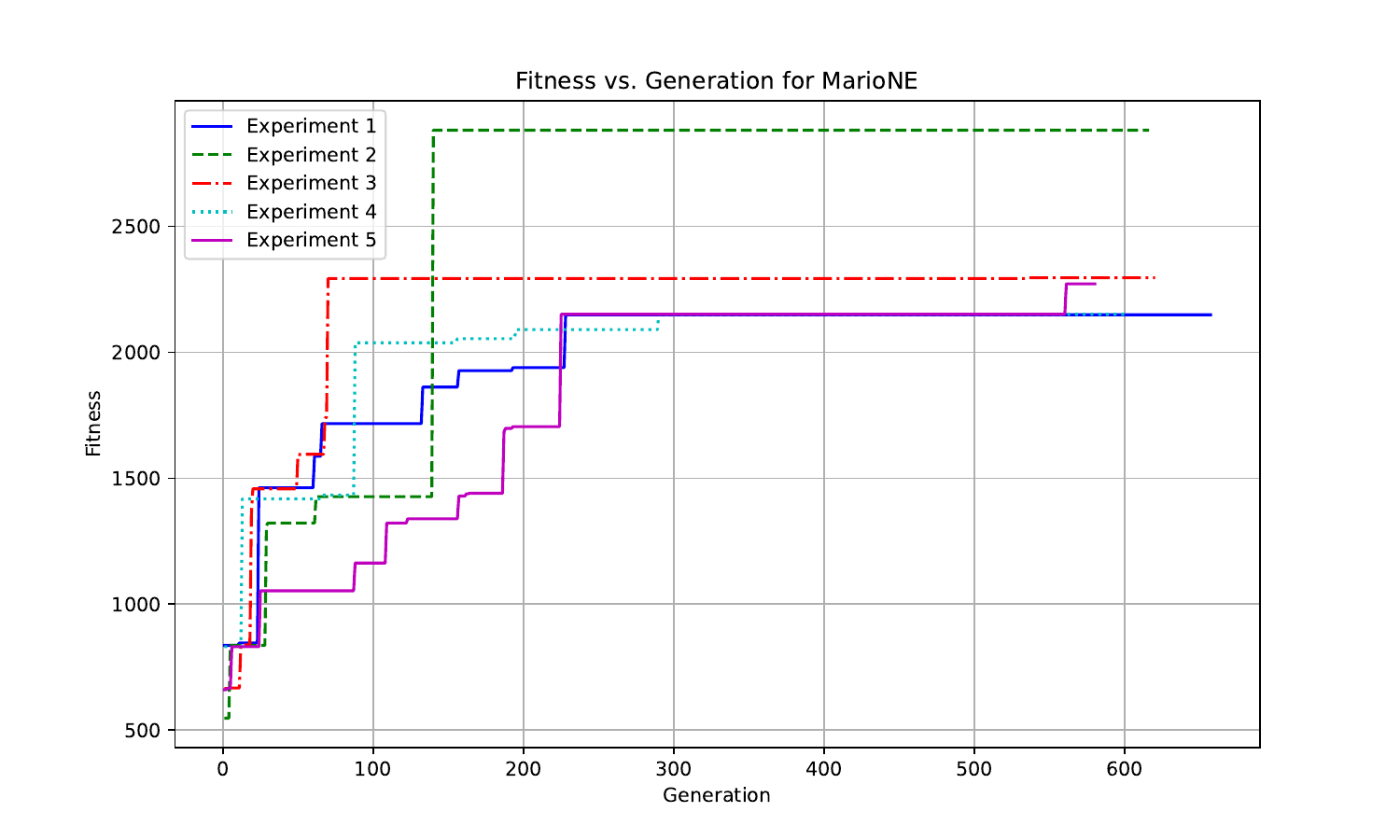}
  \caption{Fitness vs Generation plots for 800+ generations. None of the experiments were able to finish the gameplay for World 1 Level 1.
  \label{fig:Algo2plot1World11}
  }
\end{figure*}

\begin{figure*}[htp]
  \centering
  \includegraphics[width=1.99\columnwidth]{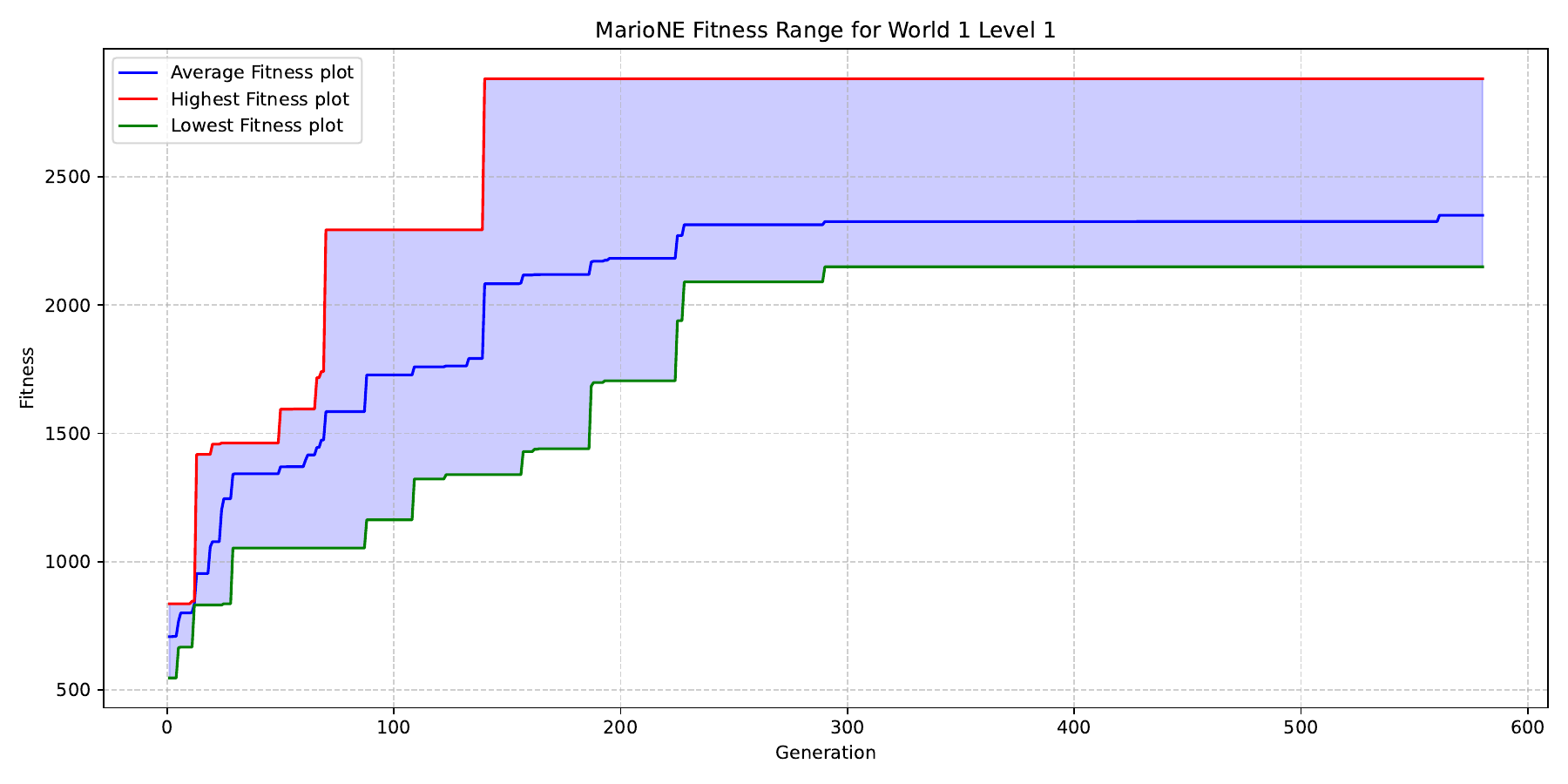}
  \caption{Highest, Average and Lowest fitness for different experiments averaged in single plot for 500+ generations.
  \label{fig:Algo2plot3World11}
  }
\end{figure*}

\begin{figure*}[htp]
  \centering
  \includegraphics[width=1.99\columnwidth]{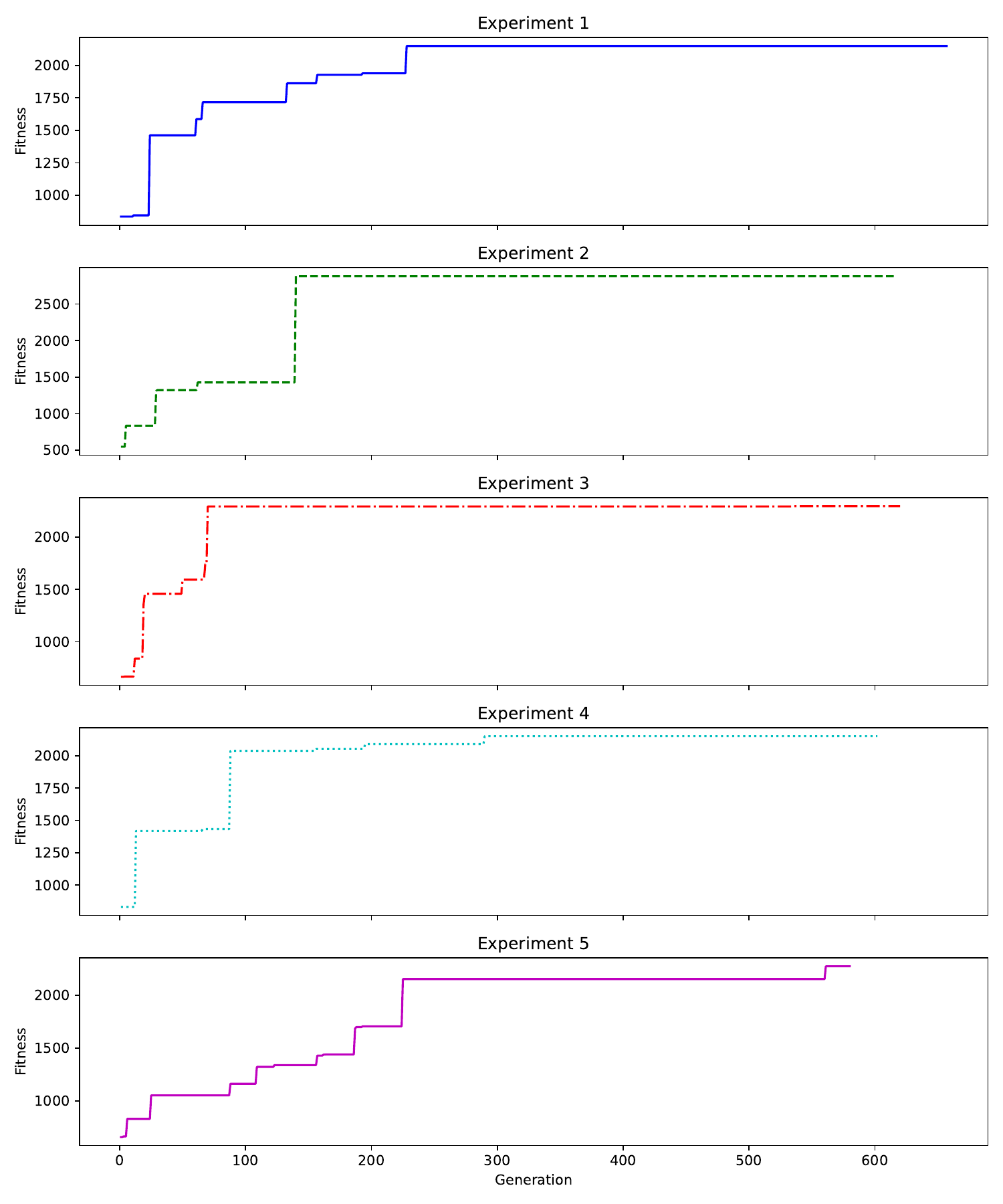}
  \caption{Multiple runs of MarioNE done for World 1 Level 1. No experiments were able to finish the experiment. 
  \label{fig:Algo2plot2World11}
  }
\end{figure*}

\begin{figure*}[htp]
  \centering
  \includegraphics[width=1.99\columnwidth]{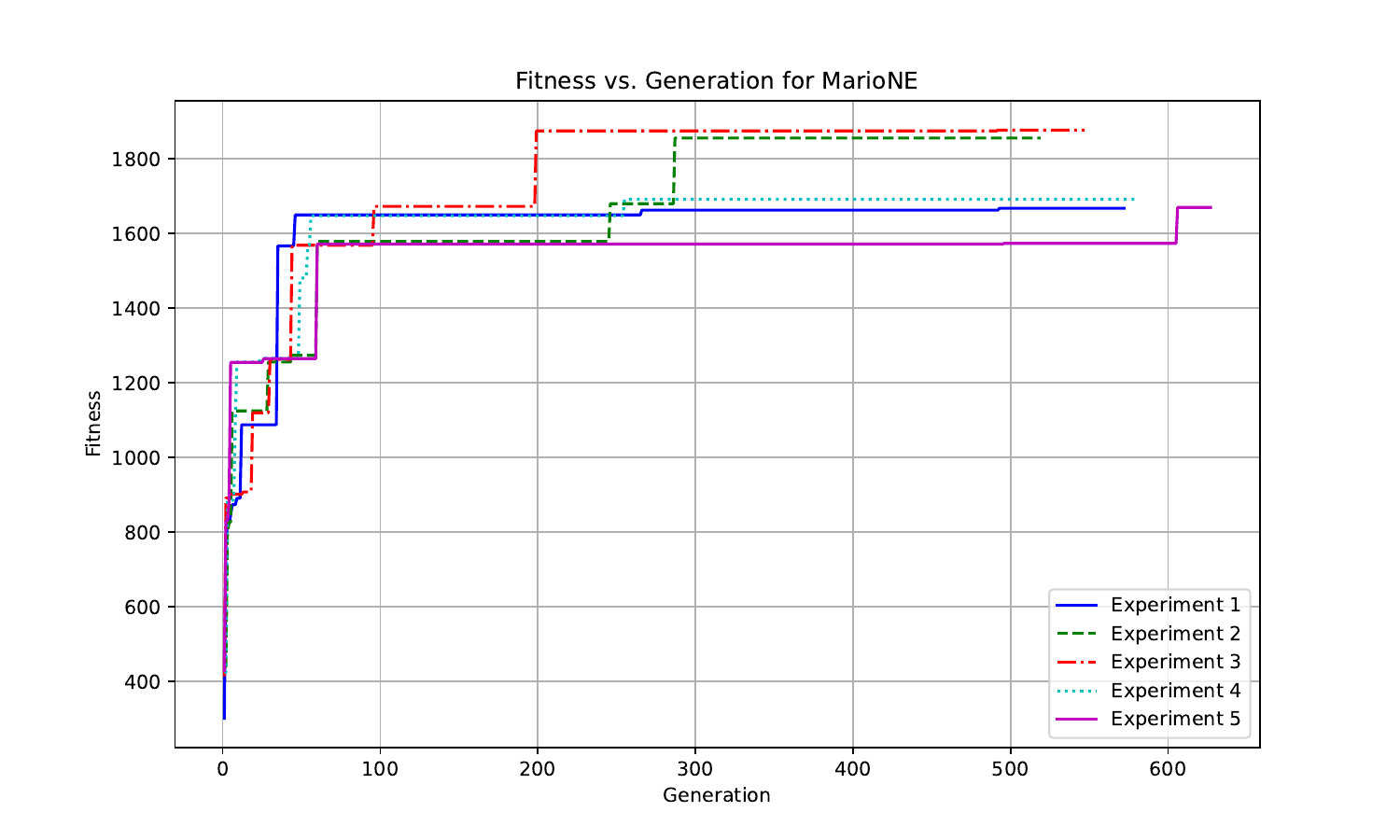}
  \caption{Fitness vs Generation plots for 800+ generations. None of the experiments were able to finish the gameplay for World 1 Level 2.
  \label{fig:Algo2plot1World12}
  }
\end{figure*}

\begin{figure*}[htp]
  \centering
  \includegraphics[width=1.99\columnwidth]{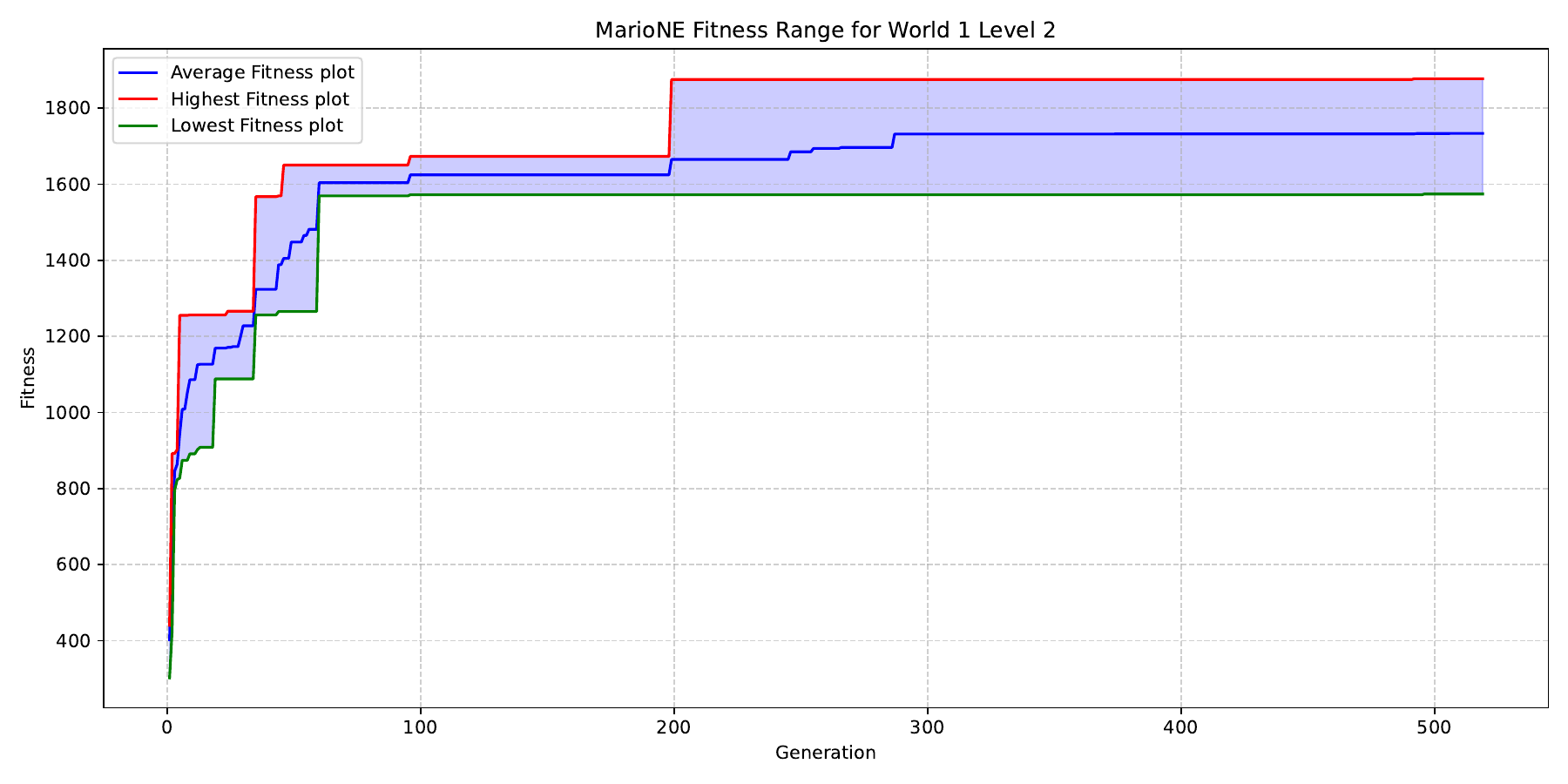}
  \caption{Highest, Average and Lowest fitness for different experiments averaged in single plot for 500+ generations.
  \label{fig:Algo2plot3World12}
  }
\end{figure*}

\begin{figure*}[htp]
  \centering
  \includegraphics[width=1.99\columnwidth]{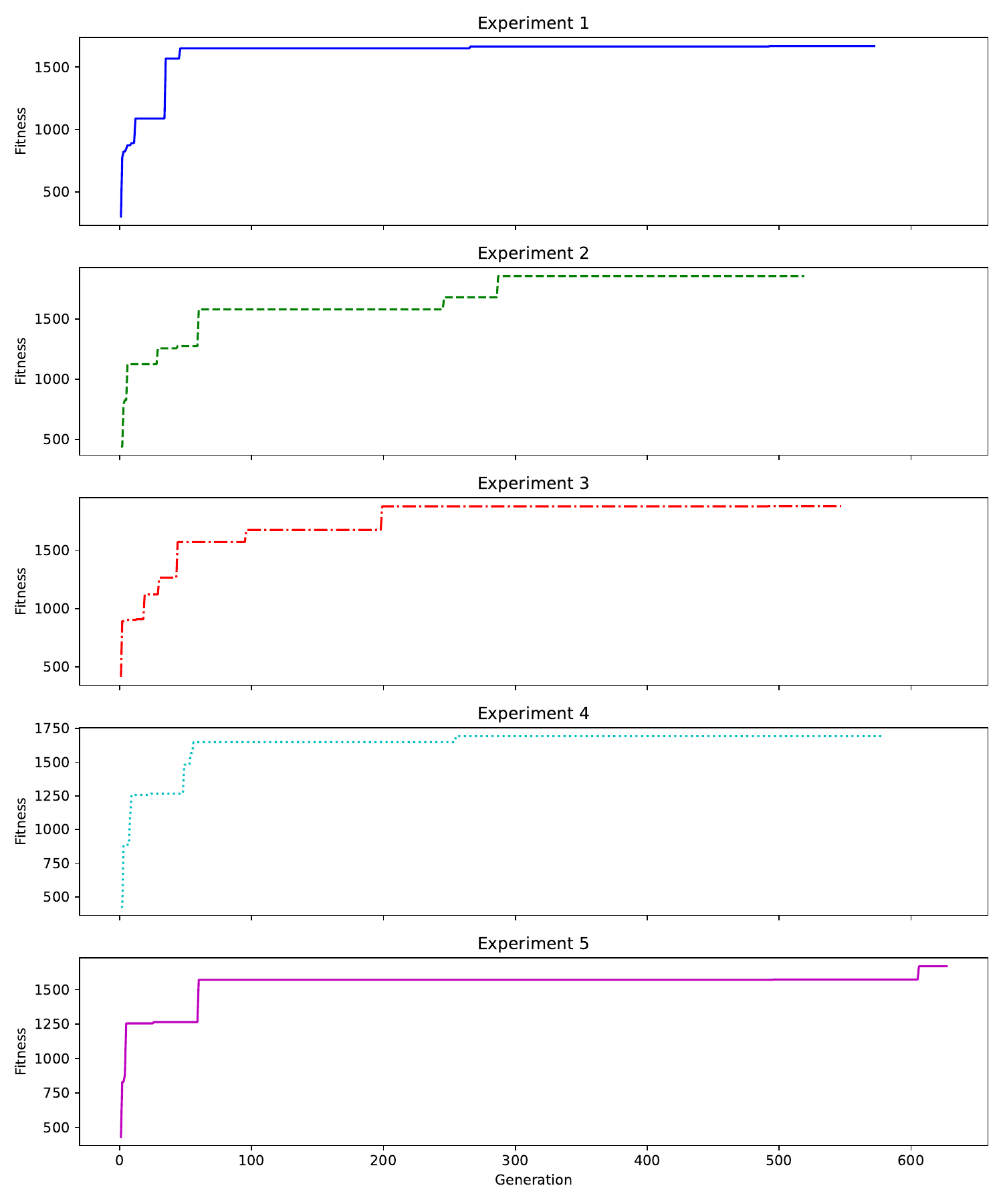}
  \caption{Multiple runs of MarioNE done for World 1 Level 2. No experiments were able to finish the experiment. 
  \label{fig:Algo2plot2World12}
  }
\end{figure*}

\begin{figure*}[htp]
  \centering
  \includegraphics[width=1.99\columnwidth]{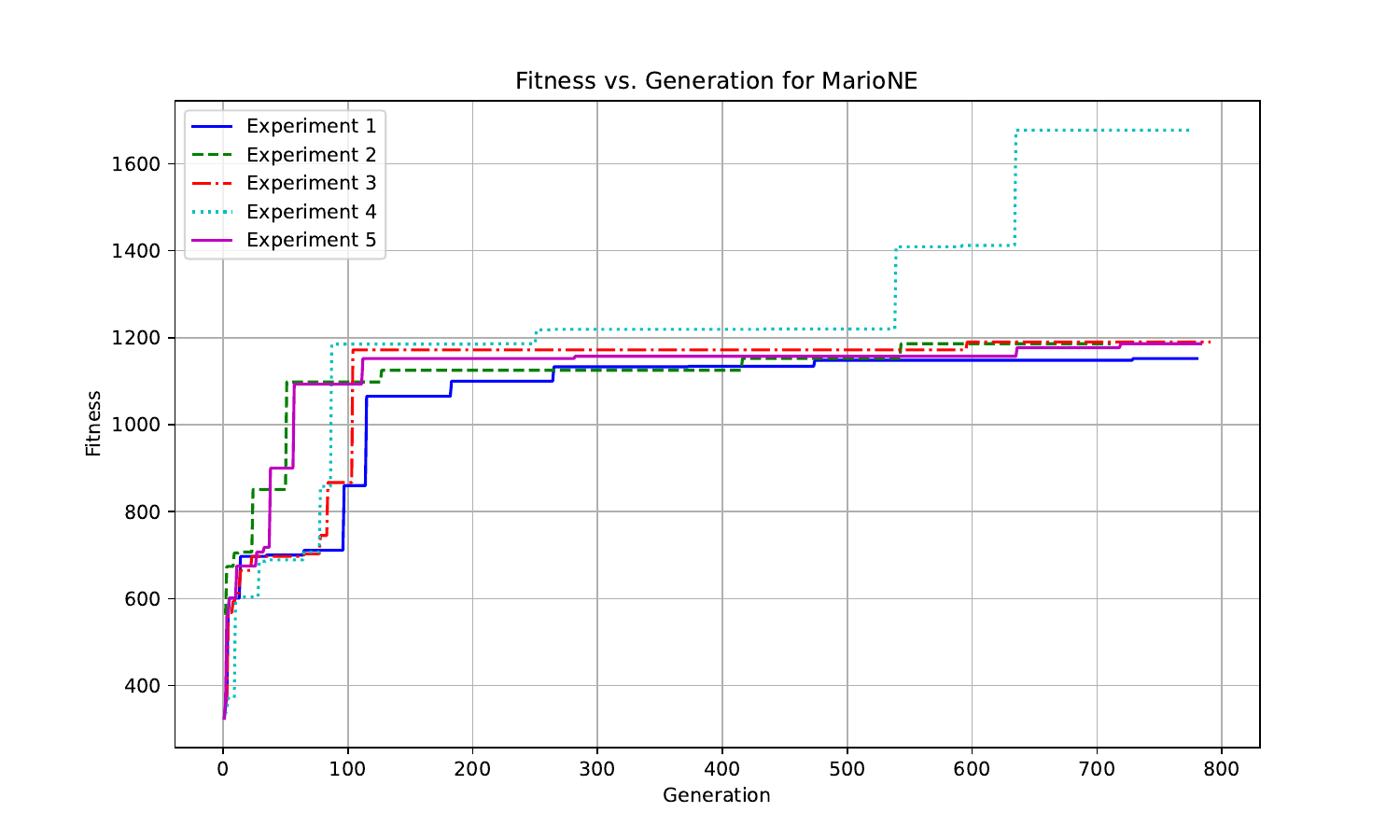}
  \caption{Fitness vs Generation plots for 800+ generations. None of the experiments were able to finish the gameplay for World 1 Level 3.
  \label{fig:Algo2plot1World13}
  }
\end{figure*}

\begin{figure*}[htp]
  \centering
  \includegraphics[width=1.99\columnwidth]{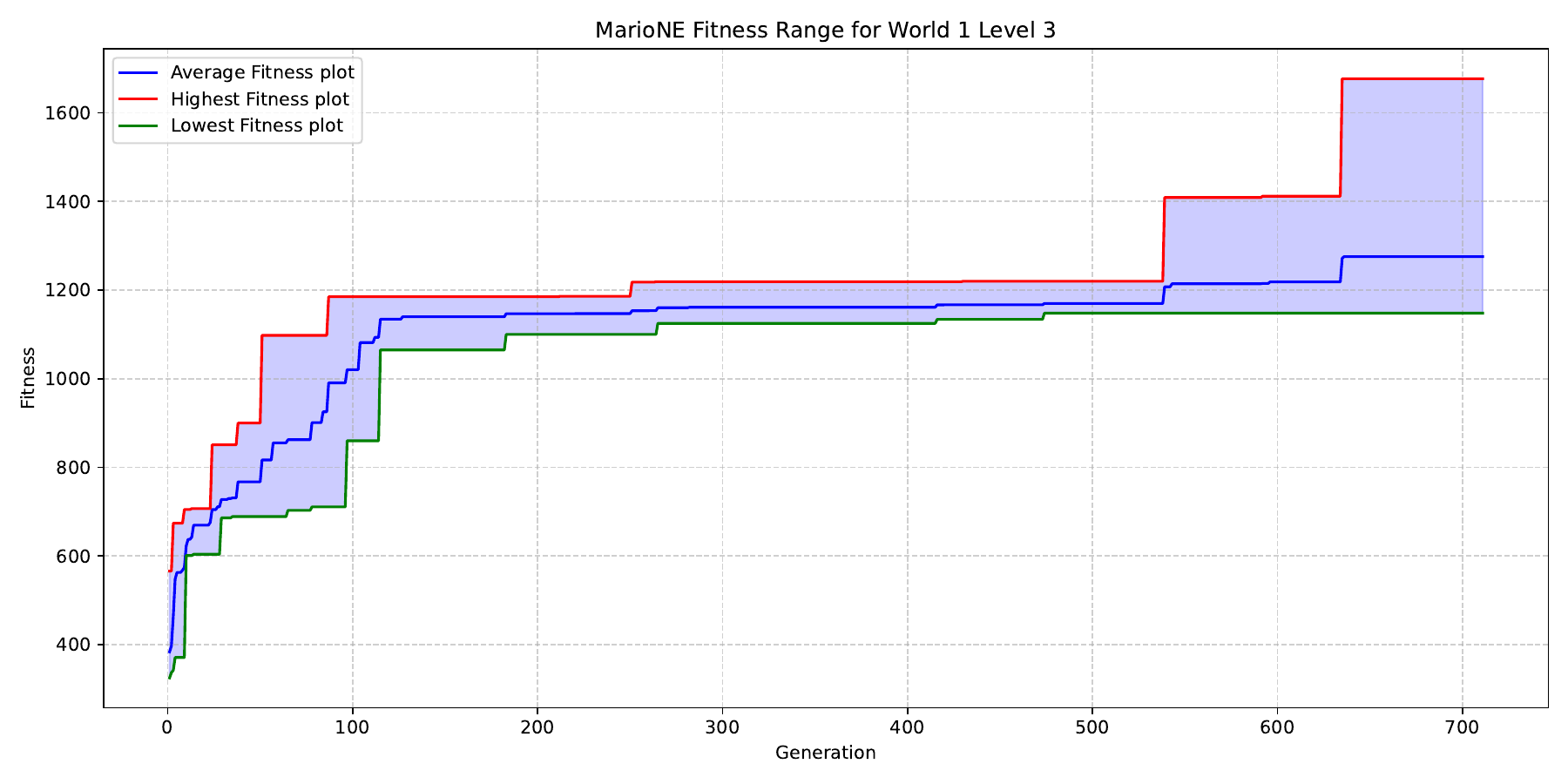}
  \caption{Highest, Average and Lowest fitness for different experiments averaged in single plot for 700+ generations.
  \label{fig:Algo2plot3World13}
  }
\end{figure*}

\begin{figure*}[htp]
  \centering
  \includegraphics[width=1.99\columnwidth]{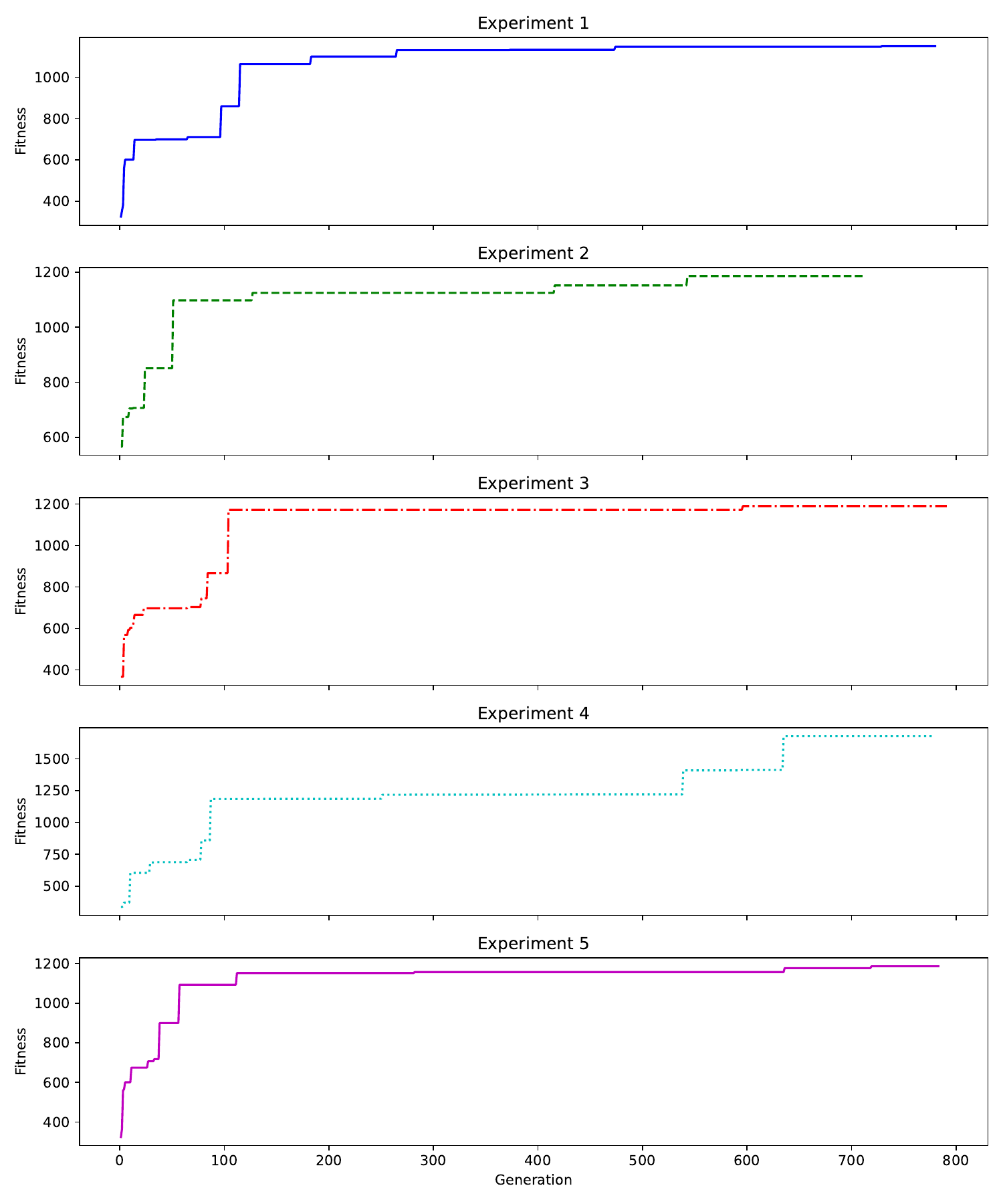}
  \caption{Multiple runs of MarioNE done for World 1 Level 3. No experiments were able to finish the experiment. 
  \label{fig:Algo2plot2World13}
  }
\end{figure*}

\begin{figure*}[htp]
  \centering
  \includegraphics[width=1.99\columnwidth]{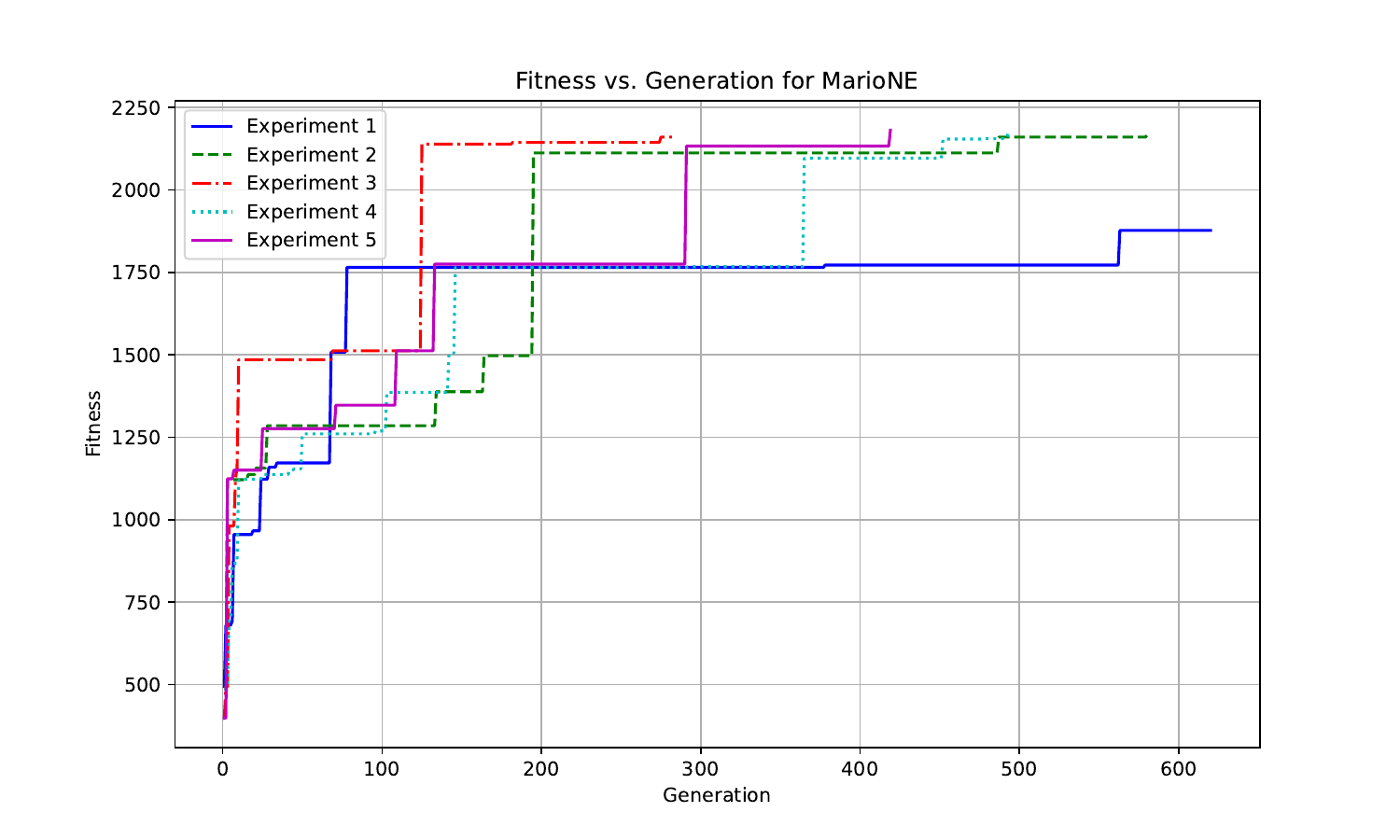}
  \caption{Fitness vs Generation plots for 500+ generations. Some experiments took so longer to converge however some of them were able to easily converge faster.
  \label{fig:Algo2plot1World14}
  }
\end{figure*}

\begin{figure*}[htp]
  \centering
  \includegraphics[width=1.99\columnwidth]{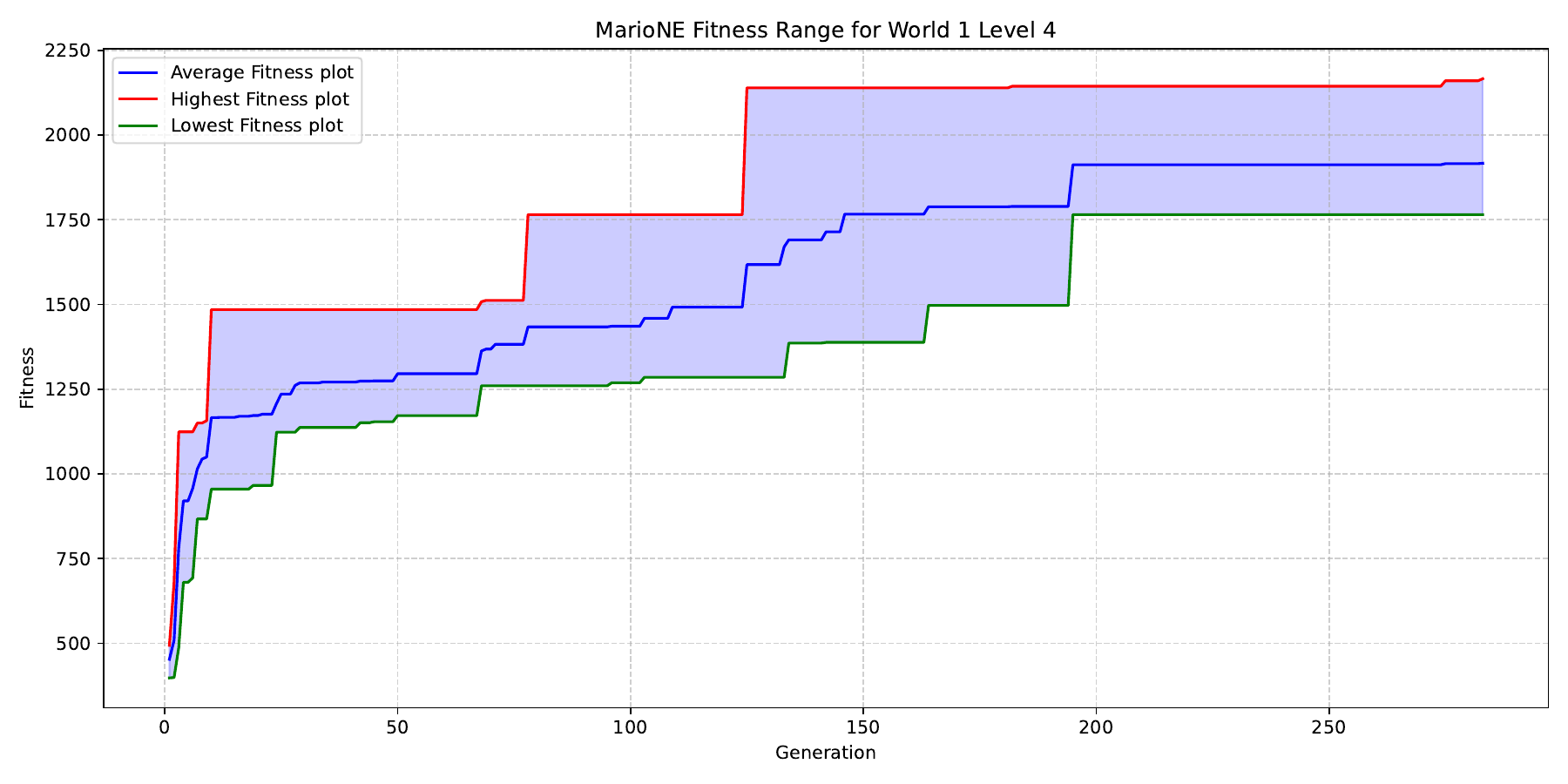}
  \caption{Highest, Average and Lowest fitness for different experiments avergaed in single plot for 280 generations generations.
  \label{fig:Algo2plot3World14}
  }
\end{figure*}

\begin{figure*}[htp]
  \centering
  \includegraphics[width=1.99\columnwidth]{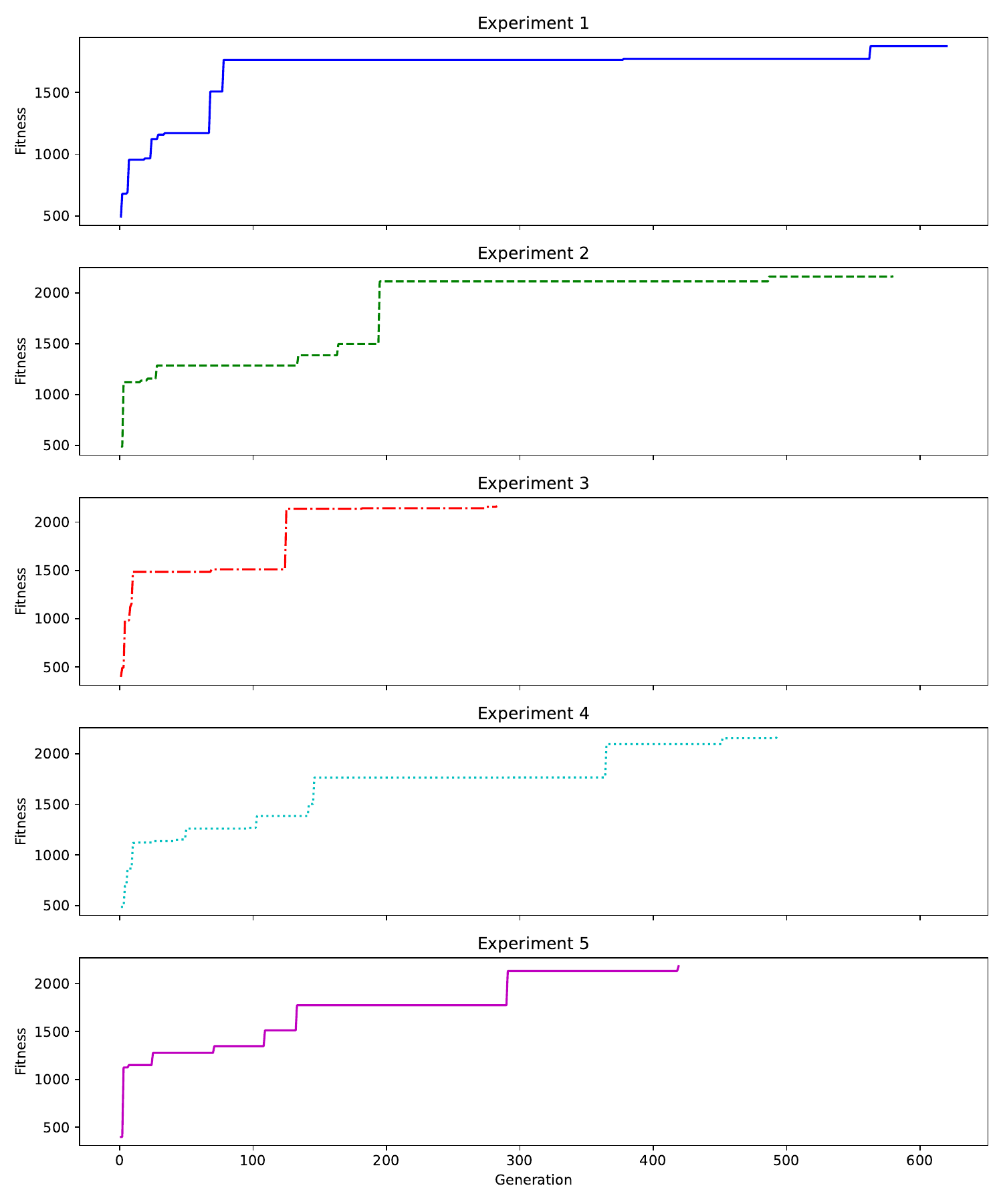}
  \caption{Multiple runs of MarioNE done for World 1 Level 4. Experiments 1, 2 and 4 were not able to finish the gameplay however Experiments 3 and 5 were able to finish it. 
  \label{fig:Algo2plot2World14}
  }
\end{figure*}

In this subsection we particularly focus on plotting Neuro Evolution experiments. We plot three figures for each world (We perform 5 experiments for each level of world one of the SMB gameplay). For details about plots we refer caption of the plot. For World 1 level 1, Figures \ref{fig:Algo2plot1World11}, \ref{fig:Algo2plot3World11}, \ref{fig:Algo2plot2World11} are shown. For World 1 level 2, Figures \ref{fig:Algo2plot1World12}, \ref{fig:Algo2plot3World12}, \ref{fig:Algo2plot2World12} are shown. For World 1 level 3, Figures \ref{fig:Algo2plot1World13}, \ref{fig:Algo2plot3World13}, \ref{fig:Algo2plot2World13} are shown. For World 1 level 4, Figures \ref{fig:Algo2plot1World14}, \ref{fig:Algo2plot3World14}, \ref{fig:Algo2plot2World14} are shown.

\textbf{Empirical comparisons} of both the algorithms are as follows:

\begin{itemize}[label=$\bullet$]
    \item \textbf{Success Rate:} MarioGA demonstrated a significantly higher success rate, successfully completing levels 1-1, 1-2, 1-3, and 1-4 in the Super Mario Bros. game with a 94\% success rate. This high success rate indicates the algorithm's effectiveness in finding successful gameplay strategies. However even after running for more than 36 hours with approx 700+ generations, MarioNE struggled to finish the gameplay. MarioNE had a lower success rate, completing levels only 8\% of the time.

    \item \textbf{Convergence Speed:} MarioGA was able to perform 8000 generations within the experimental window of 36 hours, showcasing its relatively fast convergence. The speed of convergence is an important factor when considering practical applications. MarioNE was able to perform approximately 700 generations within the experimental window of 36 hours. While this is lower than MarioGA, it still signifies a reasonable rate of convergence for neuro-evolved agents.

    \item \textbf{Average Fitness:} On average, MarioGA achieved fitness scores of 3000+ for world 1-1, 2000+ for world 1-2, 800+ for world 1-3 (which it was not able to complete and agent was stuck in local optimum in the game play as show in Figure \ref{fig:world13stuck}) and 2000+ for world 1-4, suggesting the ability to evolve highly efficient gameplay strategies. The higher average fitness indicates the algorithm's capability to optimize gameplay objectives effectively. MarioNE achieved an average fitness of 2000+ for world 1-1, 1500+ for world 1-2, 1000+ for world 1-3 and 2000+ for world 1-4, which, although lower than MarioGA, indicates the neural networks' ability to improve gameplay strategies over generations. It is essential to have a fitness of 3000+ for world 1-1, 2000+ for world 1-2, 2500+ for world 1-3 and 2500+ for world 1-4.

\begin{figure}[htp]
  \centering
  \includegraphics[width=0.99\columnwidth]{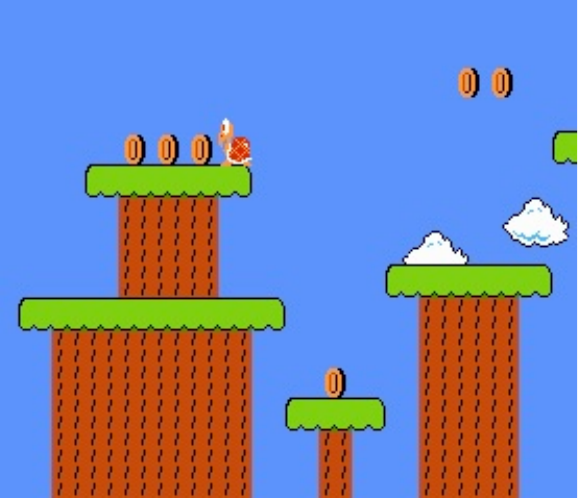}
  \caption{The SMB agent was not able to cross and jump over these pipes. 
  \label{fig:world13stuck}
  }
\end{figure}
    \item \textbf{Stability:} MarioGA demonstrated a relatively stable and consistent performance across multiple experiments. Its ability to repeatedly complete levels underscores its reliability in a Super Mario Bros. environment. MarioNE introduced the concept of evolving neural networks, offering insights into the potential of neuro-evolution. The ability to adapt neural network structures and weights is a distinct advantage.

    \item \textbf{Computational Efficiency and Running performance:} Despite a higher number of generations, MarioGA exhibited a robust computational efficiency, however, MarioNE is computationally very expensive and requires a lot of resources in order to adjust weights of the agents. While I could run 4 experiments on 4 different nodes on my CPU for MarioGA, I was only able to run 1 experiment for MarioNE on four nodes using pluralisation. The configuration of the system I used is $ml.t3.xlarge$ on Amazon Sagemaker with 500 GB storage and 16 GB of RAM.

    \item \textbf{Theoretical Time Complexity:} For MarioGA, it turns out to be $O(generation\_amount * population\_amount * moves\_amount)$, reflecting the computational cost of evolving Mario agents through generations with a fixed number of moves. On the other hand, for MarioNE, its time complexity is $O(generation\_count * population\_size * gameplay\_duration * neural\_network\_complexity)$, representing the complex interplay of evolving neural network-based Mario agents over generations, where the duration of gameplay and neural network complexity play crucial roles in computational demands.
\end{itemize}

\section{Conclusion}
In conclusion, our research has provided a comprehensive exploration of optimization strategies in the context of Super Mario Bros. gaming. MarioGA, leveraging genetic algorithms, exhibited remarkable performance, successfully completing levels 1-1, 1-2, 1-3, and 1-4 in the game with an impressive 94\% success rate. The associated fitness plots portrayed a clear evolutionary progress, revealing how MarioGA steadily improved the gameplay strategies of the Mario agents over successive generations. On the other hand, MarioNE, employing NEAT techniques, faced a more challenging path, achieving completion in only 8\% of cases. Nevertheless, this approach added a layer of complexity by evolving neural network-based Mario agents, presenting insights into the potential of NEAT in enhancing gameplay.

Our research was underpinned by meticulous five-fold cross-validation for each level, providing statistically robust results and eliminating the possibility of random luck in the performance outcomes. Furthermore, we delved into the dynamics of local optima, dataset distributions, and conducted empirical comparisons involving variations of the fitness equation constants, shedding light on the intricacies of these algorithms. The exploration of rewards and penalties within the gameplay highlighted pivotal points where significant improvements in fitness were achieved. Additionally, we conducted a rigorous analysis of the running performance of both algorithms, aligning their theoretical time complexities with their experimental results, yielding a nuanced understanding of their computational efficiency.

In sum, our research not only advances the field of AI and machine learning but also underscores the adaptability and potential of optimization strategies in complex gaming scenarios. The versatile platform provided by Super Mario Bros. continues to serve as a valuable testing ground for the development of intelligent agents, promising further exploration and refinement in both video gaming and broader AI applications.

\section{Appendix}
For the configuration of MarioNE, a standard NEAT $config$ file snapshot is shown below in table \ref{table:appendix1}:

\begin{table}
\centering
\begin{tabular}{|l|l|}
\hline
\textbf{[NEAT]} & \\
\hline
fitness\_criterion & max \\
fitness\_threshold & 500000 \\
pop\_size & 150 \\
reset\_on\_extinction & True \\
\hline
\textbf{[DefaultGenome]} & \\
\hline
activation\_default & sigmoid \\
activation\_mutate\_rate & 0.05 \\
activation\_options & sigmoid gauss \\
aggregation\_default & random \\
aggregation\_mutate\_rate & 0.05 \\
aggregation\_options & sum \\
bias\_init\_mean & 0.05 \\
bias\_init\_stdev & 1.0 \\
bias\_max\_value & 30.0 \\
bias\_min\_value & -30.0 \\
bias\_mutate\_power & 0.5 \\
bias\_mutate\_rate & 0.7 \\
bias\_replace\_rate & 0.1 \\
compatibility\_disjoint\_coefficient & 1.0 \\
compatibility\_weight\_coefficient & 0.5 \\
conn\_add\_prob & 0.5 \\
conn\_delete\_prob & 0.5 \\
enabled\_default & True \\
enabled\_mutate\_rate & 0.5 \\
feed\_forward & False \\
initial\_connection & partial 0.5 \\
node\_add\_prob & 0.5 \\
node\_delete\_prob & 0.2 \\
num\_hidden & 0 \\
num\_inputs & 960 \\
num\_outputs & 7 \\
response\_init\_mean & 1.0 \\
response\_init\_stdev & 0.05 \\
response\_max\_value & 30.0 \\
response\_min\_value & -30.0 \\
response\_mutate\_power & 0.1 \\
response\_mutate\_rate & 0.75 \\
response\_replace\_rate & 0.1 \\
weight\_init\_mean & 0.1 \\
weight\_init\_stdev & 1.0 \\
weight\_max\_value & 30 \\
weight\_min\_value & -30 \\
weight\_mutate\_power & 0.5 \\
weight\_mutate\_rate & 0.8 \\
weight\_replace\_rate & 0.1 \\
\hline
\textbf{[DefaultSpeciesSet]} & \\
\hline
compatibility\_threshold & 2.5 \\
\hline
\textbf{[DefaultStagnation]} & \\
\hline
species\_fitness\_func & max \\
max\_stagnation & 50 \\
species\_elitism & 2 \\
\hline
\textbf{[DefaultReproduction]} & \\
\hline
elitism & 3 \\
survival\_threshold & 0.3 \\
\hline
\end{tabular}
\caption{NEAT Configuration}
\label{table:appendix1}
\end{table}

\newpage
\nocite{*}
\printbibliography

\end{document}